\documentclass[letterpaper]{article} 
\usepackage{aaai2026}  
\usepackage{times}  
\usepackage{helvet}  
\usepackage{courier}  
\usepackage[hyphens]{url}  
\usepackage{graphicx} 
\usepackage{natbib}  
\usepackage{caption} 
\usepackage{algorithm}
\usepackage{algorithmic}

\usepackage[utf8]{inputenc} 
\usepackage[T1]{fontenc}    
\usepackage{url}            
\usepackage{amsmath}
\usepackage{booktabs}       
\usepackage{amsfonts}       
\usepackage{nicefrac}       
\usepackage{microtype}      
\usepackage{xcolor}    

\usepackage{newfloat}
\usepackage{listings}
\DeclareCaptionStyle{ruled}{labelfont=normalfont,labelsep=colon,strut=off} 
\floatstyle{ruled}
\newfloat{listing}{tb}{lst}{}
\floatname{listing}{Listing}
\usepackage{graphicx} 
\usepackage{amsmath}
\usepackage{bm} 
\usepackage{multirow}
\usepackage{multicol}
\usepackage{caption} 
\usepackage{enumitem}
\usepackage{subcaption} 
\usepackage{pifont}  
\usepackage{relsize}
\usepackage{colortbl}  

\usepackage{footnote} 

\usepackage{tabularx}
\usepackage{tcolorbox}

\definecolor{mygray}{gray}{.9}
\definecolor{ggray}{RGB}{127,127,127}
\definecolor{reda}{RGB}{192,0,0}
\definecolor{redb}{RGB}{217,148,143}
\definecolor{myyellow}{RGB}{190,144,0}
\definecolor{mygreen}{RGB}{80,100,40}
\definecolor{myblue}{RGB}{30,90,100}
\definecolor{myorange}{rgb}{0.9568, 0.6431, 0.3764}
\usepackage{xspace}
\makeatletter
\DeclareRobustCommand\onedot{\futurelet\@let@token\@onedot}
\def\@onedot{\ifx\@let@token.\else.\null\fi\xspace}
\def\eg{\emph{e.g}\onedot} 
\def\ie{\emph{i.e}\onedot} 
\def\cf{\emph{c.f}\onedot} 
\def\etc{\emph{etc}\onedot} \def\vs{\emph{vs}\onedot}
 
\def\etal{\emph{et al}\onedot}
\newcommand{\thickhline}{%
    \noalign {\ifnum 0=`}\fi \hrule height 1pt
    \futurelet \reserved@a \@xhline
}

\usepackage{multirow}
\usepackage{pifont}  
\usepackage{utfsym}
\usepackage{afterpage}  
\usepackage{tikz}
\usepackage[algo2e]{algorithm2e}
\usepackage{tabularx}
\usepackage{tcolorbox}

\SetKwComment{Comment}{\color{green!50!black}\# }{}

\SetKwProg{Function}{def}{:}{}

\SetKwProg{For}{for}{:}{}
\SetKwProg{If}{if}{:}{}
\newcommand{\VarSty}[1]{\textnormal{\ttfamily\color{blue!90!black}#1}\unskip}

\usepackage{algorithm}
\usepackage{algorithmic}
\usepackage{listings} 
\usepackage{colortbl}  
\usepackage{makecell}
\definecolor{codegreen}{RGB}{79,126,127}
\definecolor{codedefine}{RGB}{153,54,159}
\definecolor{codefunc}{RGB}{73,122,234}
\definecolor{codecall}{RGB}{73,122,234}
\definecolor{codepro}{RGB}{212,96,80}
\definecolor{codedim}{RGB}{89,152,195}

\definecolor{dkgreen}{rgb}{0,0.6,0}
\definecolor{gray}{rgb}{0.5,0.5,0.5}
\definecolor{mauve}{rgb}{0.58,0,0.82}

\title{Relation-R1: Progressively Cognitive Chain-of-Thought Guided Reinforcement Learning for Unified Relation Comprehension}
\author{
    Lin Li\textsuperscript{\rm 1,2}\equalcontrib, Wei Chen\textsuperscript{\rm 1}\equalcontrib,
    Jiahui Li\textsuperscript{\rm 3},
    Kwang-Ting Cheng\textsuperscript{\rm 1,2}, 
    Long Chen\textsuperscript{\rm 1}\thanks{Long Chen is the corresponding author.}
}
\affiliations{
    \textsuperscript{\rm 1} The Hong Kong University of Science and Technology \\
    \textsuperscript{\rm 2} AI Chip Center for Emerging Smart Systems \\
    \textsuperscript{\rm 3} Zhejiang University\\
    {lllidy, wchendb, timcheng, longchen}@ust.hk, jiahuil@zju.edu.cn
}

\usepackage{bibentry}

\begin{document}

\maketitle
\begin{abstract}
Recent advances in multi-modal large language models (MLLMs) have significantly improved object-level grounding and region captioning. However, they remain limited in visual relation understanding, struggling even with binary relation detection, let alone \textit{N}-ary relations involving multiple semantic roles. The core reason is the lack of modeling for \textit{structural semantic dependencies} among multi-entities, leading to over-reliance on language priors (\eg, defaulting to ``person drinks a milk'' if a person is merely holding it).
To this end, we propose Relation-R1, the \textit{first unified} relation comprehension framework that explicitly integrates cognitive chain-of-thought (CoT)-guided supervised fine-tuning (SFT) and group relative policy optimization (GRPO) within a reinforcement learning (RL) paradigm. Specifically, we first establish foundational reasoning capabilities via SFT, enforcing structured outputs with thinking processes. Then, GRPO is utilized to refine these outputs via multi-rewards optimization, prioritizing visual-semantic grounding over language-induced biases, thereby improving generalization capability. Furthermore, we investigate the impact of various CoT strategies within this framework, demonstrating that a specific-to-general progressive approach in CoT guidance further improves generalization, especially in capturing synonymous \textit{N}-ary relations. Extensive experiments on widely-used PSG and SWiG datasets demonstrate that Relation-R1 achieves state-of-the-art performance in 
both binary and \textit{N}-ary relation understanding.
\end{abstract}
\begin{links}
    \link{Code}{github.com/HKUST-LongGroup/Relation-R1}
\end{links}

\section{Introduction}

\begin{figure*}[!t]
\centering
\includegraphics[width=0.98\linewidth]{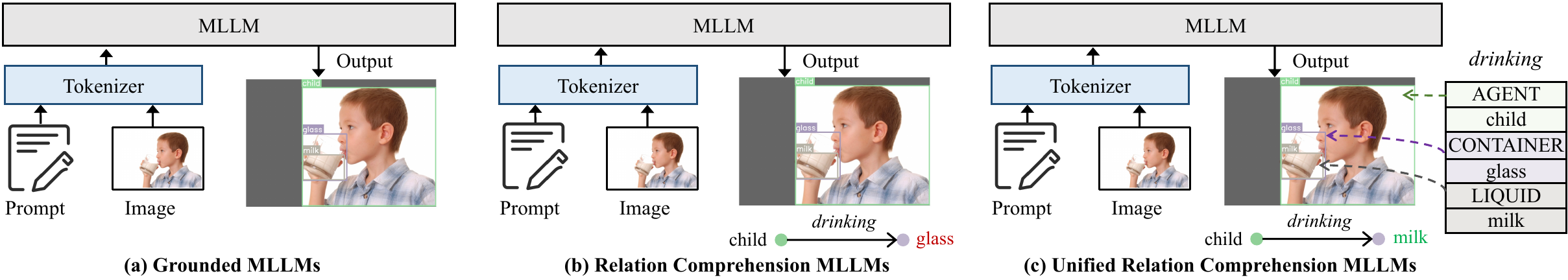}
\caption{Illustrations of our Relation-R1 and other
MLLMs. (a) \textbf{Grounded MLLMs}: Link objects in text to image regions but fail to capture object relations. (b) \textbf{Relation Comprehension MLLMs}: Model pairwise object relations but lack generalization to \textit{N}-ary interactions. (c) \textbf{Unified Relation Comprehension MLLMs}: Jointly handle binary and $N$-ary relation detection.}
\label{fig:motivation}
\end{figure*}

Recent advancements in MLLMs have significantly enhanced holistic image understanding~\cite{liu2023visual,zhu2024minigpt} and object-level grounding~\cite{youferret, zhang2024llava, rasheed2024glamm, yuan2024sr, yuan2025voxel, lai2024lisa,peng2024kosmos} capabilities, enabling tasks like region captioning and referring question answering. However, current MLLMs exhibit critical limitations in relational scene understanding -- a core competency for achieving human-like visual cognition, as illustrated in Figure~\ref{fig:motivation}(a).

Despite notable progress in binary relation detection, \ie, identifying subject-object interactions (\eg, ``\texttt{child}-\texttt{drinking}-\texttt{glass}''), current models~\cite{wang2024all,llava_spacesgg2025} struggle to capture \textit{N}-ary relationships. Accurately modeling such relationships requires models to identify multiple \textit{N}-th entities engaged in the relation with their distinct semantic roles (\textit{what}, \textit{who}, and \textit{where} concepts)~\cite{pratt2020grounded,cho2021grounded}. The neglect of these crucial \textit{structural semantic dependencies} among multi-entities, such as the glass's role as the functional \texttt{container} of milk in the drinking action, may result in suboptimal relation triplets, \eg, ``\texttt{child}-\texttt{drinking}-\texttt{glass}''. Furthermore, these unreliable and superficial pair-wise relations may make the MLLMs prone to over-reliance on language priors rather than visual-semantic cues (\eg, evidence from semantically grounded visual regions). Consequently, there arises an urgent demand for a unified framework capable of addressing these shortcomings. This raises a critical question: \textbf{\textit{How can we devise a unified framework for joint binary and \textit{N}-ary relational reasoning with grounded cues?}}

Intuitively, such complicated relation comprehension requires both strong \textbf{reasoning} and \textbf{generalization} abilities. As for \textit{reasoning}, this capability is essential as complex relations are often implicit and require the model to perform sophisticated inference by integrating diverse and often subtle visual cues. This involves a multi-step reasoning process, \eg, entity identification (\eg, recognizing ``\texttt{glass}'' category appears in the image), relation recognition (\eg, recognizing ``\texttt{drink}'' action appears in the image), role inference (\eg, ``\texttt{glass}'' as a container), localization, and contextual integration (\eg, observing a child holding a glass near their mouth suggests the ``\texttt{drink}'' action is likely occurring). In terms of \textit{generalization}, models trained on limited data struggle to extrapolate to novel interactions and entity combinations or ambiguous contexts (\eg, ``\texttt{cup}'' as a container for ``\texttt{drinking}'' action). In addition, inductive bias from language priors (\ie, common-sense assumptions learned purely from text that may not always align with the visual reality) often dominates over visual grounding, further limiting their robustness. For example, a model might struggle to recognize ``\texttt{drinking}'' from ``\texttt{milk}'' if its language training strongly associates this action with ``\texttt{glass}''.

Fortunately, DeepSeek-R1-Zero~\cite{DeepseekR1} has successfully demonstrated the emergence of reasoning capabilities in LLMs purely through RL. However, due to the inherent difficulty of reference models in understanding structural relationships, directly applying RL often struggles to produce consistently formatted outputs that are claimed in the input prompt~\cite{chu2025sft}. Conversely, when solely adopting supervised fine-tuning, the model suffers from poor generalization due to its overfitting to fixed training patterns and the limited diversity of annotated relational data.

To this end, inspired by the breakthrough of R1-series~\cite{DeepseekR1,yang2025r1onevision,zhao2025r1} in enabling reasoning and the ability of SFT to provide structured guidance, we propose a unified two-stage relation comprehension framework, \textbf{Relation-R1}, which combines the advantages of both \textbf{supervised fine-tuning} and \textbf{reinforcement learning} to empower MLLMs with relational reasoning and generalization capability:

\textbf{Stage 1. SFT}:
\label{sec:intro_sft}
Establishes foundational reasoning through step-by-step \textbf{cognitive chain-of-thought} guidance. This CoT enables the model to break down complex relation comprehension into a sequence of smaller, interpretable steps with visual evidence, such as object detection, spatial localization, \etc. Thereby, the model learns to think based on grounded visual content over language inductive bias. Furthermore, its annotated relation-contained instructions ensure that the model generates outputs in a standardized format, facilitating the subsequent computation of rewards during RL.

\textbf{Stage 2. RL}: Refines the structured outputs generated by the SFT model through group relative policy optimization. The learning process is guided by multi-rewards: 1) format rewards, which incentivize the generation of thinking process~\cite{DeepseekR1}; 2) binary relation rewards, which encourage accurate identification of pairwise relationships between entities; and 3) \textit{N}-ary relation rewards, which promote the comprehensive understanding of more complex, multi-entity activity. This stage employs policy gradient updates to cultivate the model to explore various potential solutions and optimizes its output based on defined multi-rewards, fostering robust relational reasoning and generalization.

Furthermore, to gain deeper insights into varying cognitive CoTs, we investigate the impact of template-based CoT and MLLM-generated CoT strategies on binary and \textit{N}-ary relation comprehension, separately. Notably, by employing a progressive paradigm that initially adopts the template-based CoTs to guide the normative cognitive reasoning process, and then fine-tunes the model with the variable reasoning pathways introduced by a few MLLM-generated CoTs, our Relation-R1 demonstrates better generalization ability. Especially in the \textit{N}-ary relations, the capacity to explore \textit{synonymous relational expressions} has emerged.

We validate the effectiveness of Relation-R1 on both binary relation comprehension (\eg, scene graph generation~\cite{xu2017scene,wang2024all}) and \textit{N}-ary relation comprehension task (\eg, grounded situation recognition~\cite{pratt2020grounded}) across diverse experimental settings. Experiments on the PSG~\cite{yang2022panoptic} and SWiG~\cite{pratt2020grounded} datasets demonstrate state-of-the-art performance in relational reasoning while maintaining generalization in ambiguous contexts.
In summary, our contributions are fourfold:
\begin{itemize}[itemsep=0pt, parsep=0pt, topsep=0pt, partopsep=0pt, leftmargin=20pt]
\item We reveal the limitations of the existing MLLMs in relation understanding, and incorporate both binary and \textit{N}-ary relation detection into one unified relation comprehension framework.
\item We propose Relation-R1, a unified framework that integrates cognitive CoT-guided SFT with RL. To the best of our knowledge, this is the first work that enables both relational reasoning and robust generalization in a deep-thinking fashion.
\item We comparatively analyze template-based and MLLM-generated CoTs, demonstrating that progressively guiding the learning process with CoTs in a specific-to-general manner enhances generalization capabilities, particularly in capturing synonymous \textit{N}-ary relations.
\item  Experimental results on the prevalent relation understanding datasets (\eg, +\textbf{6.84}$\sim$\textbf{6.90}\% gains on PSG dataset) under different settings demonstrate the effectiveness of Relation-R1.
\end{itemize}

\section{Related Work}
\textbf{Reasoning in MLLMs.}
Recent LLMs have demonstrated advanced reasoning capabilities by emulating human-like stepwise thinking, significantly enhancing performance on complex tasks~\cite{jaech2024openai,DeepseekR1}. A pivotal breakthrough, DeepSeek-R1~\cite{DeepseekR1}, leverages large-scale RL with formatting and result-oriented rewards to enable LLMs to autonomously generate human-like complex Chain-of-Thought (CoT) reasoning, achieving state-of-the-art results across diverse domains. MLLMs extend these capabilities through systematic integration of cross-modal knowledge representation and task-specific rewards~\cite{yang2025r1onevision, du2025virgo, huang2025vision, shen2025tarpro}
For multimodal mathematical
reasoning, frameworks such as R1-Onevision~\cite{yang2025r1onevision}, Virgo~\cite{du2025virgo}, and MM-Eureka~\cite{meng2025mm} jointly reason over visual and textual inputs to solve quantitative tasks, while Vision-R1~\cite{huang2025vision} combines cold-start initialization with GRPO to refine complex CoT reasoning.  For fine-grained classification and grounding tasks, Visual-RFT~\cite{liu2025visual} and R1-Omni~\cite{zhao2025r1} utilize GRPO-based reinforcement algorithms and verifiable rewards to enhance contextual reasoning. As for the pixel-level understanding task, SegZero~\cite{liu2025seg} employs reinforcement learning to achieve high-resolution semantic segmentation. Despite these advances, existing MLLMs predominantly focus on \textit{object-centric recognition} rather than unified visual relation understanding, which requires compositional reasoning about geometric, semantic, and functional interactions between objects and contexts. To bridge this gap, Relation-R1 explicitly integrates \textit{relational reasoning} into MLLMs via RL, enabling systematic inference for both scene interpretation and event prediction.

\noindent\textbf{Scene Graph Generation (SGG).} SGG is a critical task in scene understanding, with prior approaches falling into two groups: 1) \textit{Two-stage SGG}, which sequentially detects objects and infers pairwise relations but suffers from error propagation~\cite{tang2020unbiased,zellers2018neural,li2022devil,li2023compositional,shi2025easy}; 2) \textit{One-stage SGG},
which unifies detection and relation prediction in end-to-end frameworks (\eg, DETR-based methods~\cite{carion2020end,li2022sgtr,chen2020simple}) but lacks multi-role involved relation modeling. Recent efforts in \textit{open-vocabulary SGG (OVSGG)} employ Vision-Language Models (VLMs)~\cite{chen2024scene,li2023zero,li2025inter,chen2024expanding} or MLLMs~\cite{wang2024all,llava_spacesgg2025,li2024pixels} to handle novel entities/relations, yet limited modeling of complex multi-entity interactions, and suffer from overfitting in base categories during SFT. Our Relation-R1 addresses OVSGG challenges by adopting a cognitive CoT-guided RL framework to model complex multi-entity interactions and mitigate SFT overfitting, enabling robust zero-shot reasoning without predefined category constraints.

\noindent\textbf{Grounded Situation Recognition (GSR).} 
GSR extends scene understanding by jointly modeling actions (verbs) and their associated semantic roles. Pratt \etal~\cite{pratt2020grounded} pioneered this field with a two-stage RNN-based framework: verb detection followed by noun localization. Subsequent works~\cite{cho2021grounded,wei2021rethinking,cho2022collaborative,cheng2022gsrformer} improved this pipeline by integrating transformers and semantic relation modeling, enabling more coherent scene interpretations. Despite these efforts, previous efforts that depend on annotated training samples may face challenges like long-tailed distribution~\cite{tang2020unbiased,li2024nicest,shao2024knowledge,chen2023addressing}, potentially limiting real-world applicability. While LEX~\cite{lei2024seeing} realizes training-free zero-shot GSR, it fails to provide an end-to-end framework and exhibits suboptimal performance and computational efficiency due to its reliance on LLMs' generated descriptions. This paper proposes Relation-R1, a unified framework that bridges the gap between open-ended capabilities and end-to-end training.

\begin{figure*}[t]
\centering
    \includegraphics[width=0.95\linewidth]{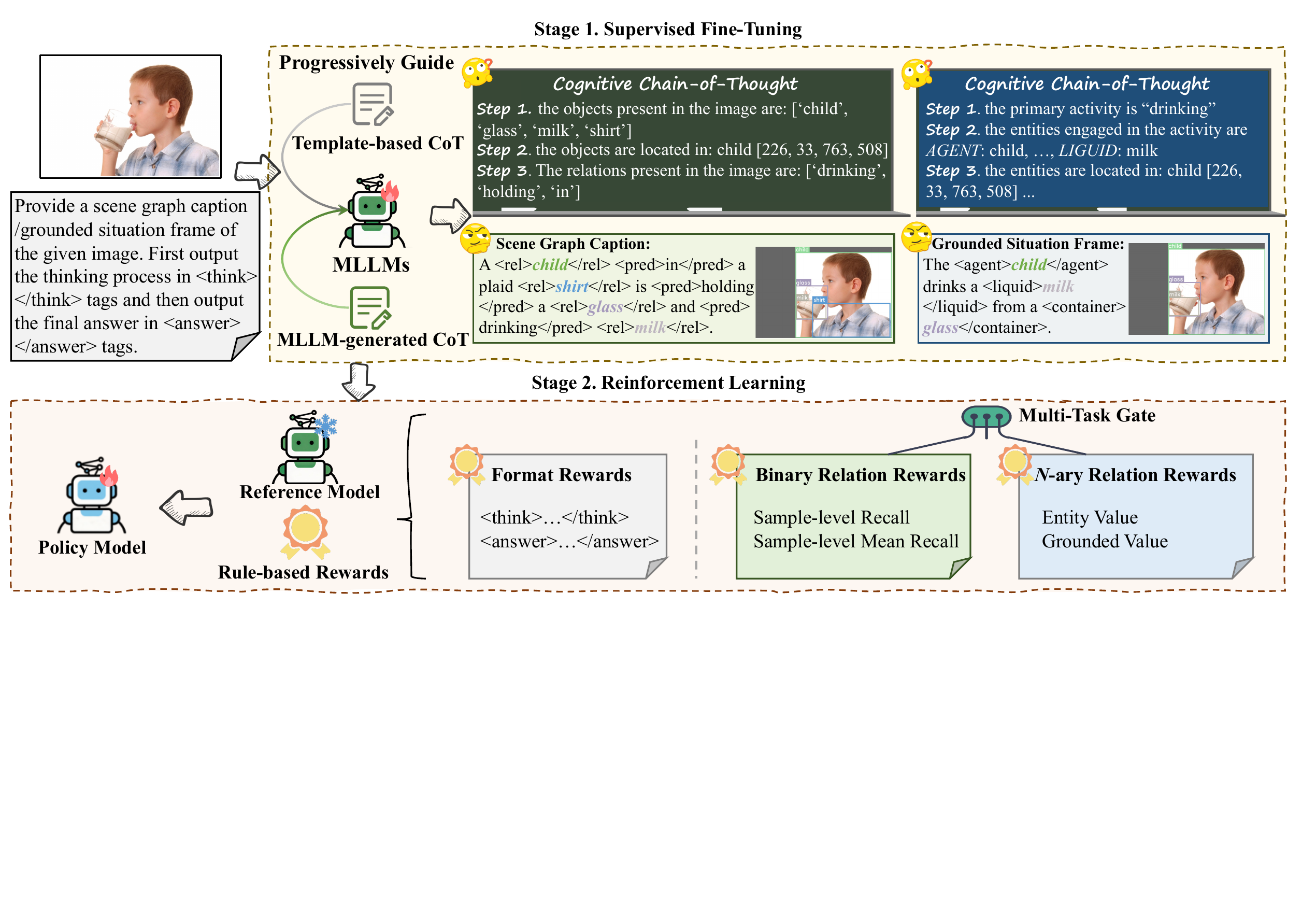}
     \caption{\textbf{Relation-R1} framework. (a) \textbf{SFT} with progressive guidance of cognitive CoT, which enforces task-specific reasoning processes and format alignment. (b) \textbf{RL} with GRPO, employs rule-based rewards (\eg, format rewards, binary relation rewards, and \textit{N}-ary relation rewards) to cultivate robust relation reasoning and generalization ability through policy gradient updates.}
    \label{fig:overview}
\end{figure*}

\section{Methodology}

\noindent\textbf{Preliminaries.} 1)
\textit{Binary Relation Detection}: This task requires the model to generate an \textit{open-ended scene graph caption} given an image, explicitly marking semantic relationships between pairwise objects~\cite{wang2024all}. The output is a triplet-contained caption annotated with formal tags: objects are denoted by <\texttt{ref}> tags followed by their bounding box coordinates (top-left/bottom-right) in <
\texttt{box}>, \eg, <\texttt{ref}>\textit{person}</\texttt{ref}><\texttt{box}>[[$x_1$, $y_1$, $x_2$, $y_2$]]</\texttt{box}>. Predicates are enclosed in <\texttt{pred}> tags, followed by references to their \textit{subject} and \textit{object} entities via their corresponding bounding boxes. 2) \textit{\textit{N}-ary Relation Detection:} It aims to output a \textit{open-ended grounded situation frame}~\cite{pratt2020grounded} structured as a primary activity (\ie, multi-entity involved relation), followed by a sequence of <\texttt{role}> tags along with corresponding bounding box coordinates. Each role explicitly defines the entity's participation in the activity, \eg, <\texttt{agent}>\textit{person}</\texttt{agent}><\texttt{box}>[$x_1$, $y_1$, $x_2$, $y_2$]</\texttt{box}>. 

\subsection{Supervised Fine-Tuning}
\label{sec:method_sft}
As discussed above, directly applying GRPO encounters challenges in generating consistently formatted scene graph captions and grounded situation frame outputs. To address this, we employ SFT as a foundational stage to establish task-specific output format adherence (\cf, Figure~\ref{fig:overview}(a)).

However, relying solely on the target answer format during SFT risks compromising the model's underlying reasoning capabilities by encouraging a superficial alignment with formulated structures~\cite{chu2025sft}. To mitigate this potential issue, we introduce the \textbf{cognitive chain-of-thought} that integrates stepwise cognitive processes such as object classification, object grounding, and visual relationship inference. These cognitive CoTs are explicitly encapsulated within <\texttt{think}> tags, during SFT. This ensures that the model retains its capacity for multi-step reasoning while learning to produce structured outputs conforming to the desired format.

To streamline the think process and prevent the MLLM from overfitting to specific CoT patterns during SFT, we adopt a specific-to-general progressive training strategy:

\noindent\textbf{Template-based CoT.} For the specific learning phase, we devise a fixed, template-based CoT. As illustrated in Figure~\ref{fig:overview}(a), this template comprises a predefined sequence of steps for both binary and \textit{N}-ary relation detection. To elaborate, for binary relation detection, the template includes: 1) Object Existence (\eg, \texttt{the objects present ...}); 2) Object Localization (\eg, \texttt{the objects are located ...}); and 3) Relation Existence (\eg, \texttt{the relations present ...}). For \textit{N}-ary relation detection, it involves: 1) Activity Recognition (\eg, \texttt{the primary activity is ...}); 2) Entities and Roles Recognition (\eg, \texttt{the entities engaged in the activity are...}); and 3) Entity Localization (\eg, \texttt{the entities are located in ...}).

\noindent\textbf{MLLM-generated CoT.} For the general reasoning process, we design a \textbf{cognitive CoT generation prompt} that explicitly makes an extra strong MLLM (\eg, Qwen 2.5-VL~\cite{Qwen2.5-VL}) to generate step-by-step reasoning processes. To be specific, this cognitive CoT generation prompt consists of a task definition, a ground-truth (GT) scene graph of the current image, and a CoT generation instruction. The former one makes the MLLM generate the correct task format. Meanwhile, the latter two anchors reasoning content to visual evidence, ensuring that the MLLM's cognitive CoT aligns with both semantic validity and visual grounding rather than relying on category priors. The detailed cognitive CoT generation prompt is left in the Appendix. 

\noindent\textbf{Progressive CoT Guidance.} With a progressive paradigm that involves initial SFT training on easy-to-learn template-based CoTs and subsequent fine-tuning on a small number of MLLM-generated CoTs, the MLLM gains foundational capability to generate format-compliant outputs and grounded reasoning processes, contributing to GRPO training.

\subsection{Group Relative Policy Optimization}
Following the Deepseek-R1~\cite{DeepseekR1}, the group relative policy optimization is adopted as the RL algorithm for optimizing our Relation-R1 framework, as illustrated in Figure~\ref{fig:overview}(b). Distinct from critic-dependent methods~\cite{schulman2017proximal}, GRPO directly calculates advantage estimates through response group comparisons sampled from the policy model, inherently eliminating the need for an extra critic network and reducing computational complexity~\cite{DeepseekR1}. Key components include the a question $q$, policy model $\pi_{\theta}$, response group $\{o_1, o_2, \ldots, o_G \}$, and a frozen reference model $\pi_{\theta_{\text{ref}}}$, which acts as a regularization baseline to preserve prior knowledge. The GRPO objective function $J_{\text{GRPO}}(\theta)$ balances reward maximization and policy stability by leveraging the following objectives:
\begin{equation}
\small
\begin{aligned}
&J_{\text{GRPO}}(\theta) = \mathbb{E}_{q \sim \mathcal{Q}, \{o_i\}_{i=1}^G \sim \pi_{\theta_{\text{old}}}}  \\
&[\frac{1}{G}\!\sum_{i=1}^{G} \!\min\left(\!\rho_i A_i,\text{clip}(\rho_i,1\!-\!\epsilon, 1\!+\!\epsilon) A_i\!\right)\!-\!\beta\!D_{\text{KL}}(\!\pi_{\theta}\!\|\!\pi_{\text{ref}})],
\label{eq:grpo}
\end{aligned}
\end{equation}
where $\rho_i$ = $\frac{\pi_{\theta}(o_i|q)}{\pi_{\theta_{\text{old}}}(o_i|q)}$ quantifies policy change, $\epsilon$ controls clipping thresholds, and $\beta$ penalizes deviations via the KL divergence term. The advantage score $A_i$ is standardized as:
\begin{equation}
\small
A_i = \frac{r_i - \text{mean}\left( \{ r_1, \ldots, r_G \} \right)}{\text{std}\left( \{ r_1, \ldots, r_G \} \right)},
\label{eq:advantage}
\end{equation}
here $r_i$ denotes the reward for response $o_i$. The KL divergence term enforces proximity to the reference policy:
\begin{equation}
\small
D_{KL}(\pi_\theta \| \pi_{\text{ref}}) = \frac{\pi_{\text{ref}}(o_i | q)}{\pi_\theta(o_i | q)}- \log \left( \frac{\pi_{\text{ref}}(o_i | q)}{\pi_\theta(o_i | q)} \right) - 1.
\label{eq:kl}
\end{equation}
This term ensures controlled exploration without excessive divergence from $\pi_{\theta_{\text{ref}}}$.

The reward $r_i$ is formulated to align with the model's learning objective for the unified relation understanding. It incorporates rule-based rewards, which consist of three crucial components: Format compliance rewards, binary relation rewards, and \textit{N}-ary relation rewards. The former enforces syntactic validity by penalizing outputs that deviate from predefined structural constraints. The latter two components focus on task-specific performance.

\noindent\textbf{Format Rewards.} This reward enforces strict adherence to the reasoning template format: <\texttt{think}> \textit{thinking process} </\texttt{think}><\texttt{answer}> \textit{answer} </\texttt{answer}>.
The reward function is defined as:  
\begin{equation}
\small
r_{\text{form}}(o_i) = 
\begin{cases} 
1 & \text{if } o_i \text{ adheres to the format}, \\
0 & \text{otherwise}.
\end{cases}
\label{eq:template_reward}
\end{equation}  
This binary mechanism ensures syntactic compliance while allowing content flexibility.

\noindent\textbf{Binary Relation Rewards.} Inspired by established evaluation practices~\cite{yang2022panoptic,wang2024all}, we devise the sample-level triplet recall $R$ and sample-level mean recall $mR$ in the response of each sample. The former is the ratio of correctly predicted triplets to the total ground-truth triplets in a sample; the latter is the average recall across all predicate categories.
Specifically, a triplet is deemed correct if:
\begin{itemize}[itemsep=0pt, parsep=0pt, topsep=0pt, partopsep=0pt, leftmargin=10pt]
    \item \textbf{Triplet Accuracy}: All entity and relation categories (\texttt{subject}, \texttt{predicate}, \texttt{object}) in this sample are correctly identified.
    \item \textbf{Spatial Localization}: The predicted bounding boxes for the subject and object achieve an Intersection over Union (IoU) >= 0.5 with their respective ground-truth boxes.
\end{itemize}
The final binary relation reward is computed as a weighted combination of these two metrics: 
\begin{equation}
\small
    r_{\text{binary}}(o) = \alpha \cdot R + (1-\alpha) \cdot mR,
\label{eq:binary_reward}
\end{equation}
where $\alpha$ balances the trade-off between two metrics. 

\noindent\textbf{\textit{N}-ary Relation Rewards.}
Similarly, to test higher-order relational structures in grounded situation frames, we devise \textit{N}-ary relation rewards to assess the model's ability to detect primary activity (verb) and multi-entity interactions. Specifically, we calculate the entity value $V_e$ and grounded value $V_{grnd}$ for each sample~\cite{pratt2020grounded}. They represent the correct proportion in predicted entity classes and roles, and the spatial localization of those entities, respectively:
\begin{itemize}[itemsep=0pt, parsep=0pt, topsep=0pt, partopsep=0pt, leftmargin=10pt]
    \item \textbf{Entity Accuracy}: An entity within an event is considered correct if its predicted category and semantic role align with the GT, and the predicted activity matches the GT.
    \item \textbf{Spatial Localization}: The predicted bounding box of each role obtains IoU >=0.5 with the GT's bounding box.
\end{itemize}
The eventual \textit{N}-ary relation reward is computed as a weighted combination of these two accuracies: 
\begin{equation}
\small
    r_{\text{n-ary}}(o) = \beta \cdot V_e + (1-\beta) \cdot V_{grnd},
\label{eq:nnary_reward}
\end{equation}
where the hyperparameter $\beta$ balances the trade-off between the two values. As the reward's calculation depends on accurate verb prediction, they jointly evaluate primary activity and multi-entity interactions. During training, we design a multi-task gate that dynamically selects task-specific rewards based on the presence of <\texttt{ref}> tags in the solutions.

\section{Experiment}
\label{sec:exp}
{\textbf{Datasets.}} 1) \textbf{Panoptic Scene Graph (PSG)}: The official PSG dataset~\cite{yang2022panoptic} comprises 48,749 annotated images of which 46,563 are for training and 2,186 are for testing. It integrates 80 ``thing'' object categories and 53 ``stuff'', alongside 56 relation categories. Notably, ASMv2~\cite{wang2024all} reconstructs the dataset into a scene graph caption format, utilizing a split of 42,250 training images and 1,000 evaluation images. To mark fair comparison, we followed ASMv2~\cite{wang2024all}, using scene graph caption where objects are tagged with <\texttt{ref}> and paired with bounding box coordinates specifying their locations, while predicates are marked with <\texttt{pred}> and linked to two bounding boxes indicating their subject and object entities for training. Besides, we also present results using the conventional scene graph format, \ie, a list of visual triplets structured as [subject, subject bounding box, object, object bounding box, relation] on the official dataset\footnote{More details of the scene graph format are in the Appendix\label{footnote:sgg_format}.}.
2) \textbf{SWiG}: The SWiG dataset~\cite{pratt2020grounded} extends the imSitu dataset~\cite{yatskar2016situation} by adding extra bounding box annotations for semantic roles, with 69.3\% of semantic roles localized. Each image is annotated with a verb and a variable number of semantic roles (1$\sim$6 roles per image), where each verb is associated with three verb frames annotated by independent human annotators. The dataset comprises 25,200 testing images spanning 504 verb categories, 190 semantic role categories, and 9,929 noun entity categories. Unlike traditional output structures that merely predict categories and bounding boxes, we constructed grounded situation frames by filling verb-centric templates (\eg, ``The AGENT drinks a LIQUID from a CONTAINER'') with three critical components: role tags (\eg, <\texttt{agent}>, entity categories (\eg, child), and coordinates of corresponding bounding boxes in <\texttt{box}> tags.

\begin{table}[!t]
    \centering
    \small
    \renewcommand{\arraystretch}{1.0}
    \begin{tabular}{|r||c|c|c|c|}
    \thickhline
    \rowcolor{mygray}
    \!Method\! & \!Size\! & \!Recall\! & \!mRecall\! & \!Mean\! \\
    \hline
    \multicolumn{5}{|c|}{\textit{Close-ended}} \\
    \hline
\!IMP\scriptsize{~\cite{xu2017scene}}\!  & \!-\! & \!16.50\! & \!6.50\! & \!11.50\! \\
    \!MOTIFS\scriptsize{~\cite{zellers2018neural}}\! & \!-\! & \!20.00\! & \!9.10\! & \!14.55\!\\
    \!VCTree\scriptsize{~\cite{tang2019learning}}\!  &  \!-\!  & \!20.60\! & \!9.70\! & \!15.15\!\\
    \!GPSNet\scriptsize{~\cite{lin2020gps}}\! &  \!-\!  &\!17.80\! & \!7.00\! & \!12.4\! \\
\!PSGFormer\scriptsize{~\cite{yang2022panoptic}}\! &  \!-\! &  \!18.60\! & \!16.70\! & \!17.65\!\\
  \hline
  \multicolumn{5}{|c|}{\textit{Open-ended}} \\
  \hline
    \!TextPSG$^\star$\scriptsize{~\cite{Zhao_2023_ICCV}}\!  &  \!-\!  & \!4.80\! & \!-\! & \!-\!\\
    \!R1-SGG$^\star$\scriptsize{~\cite{chen2025compile}}\!  & \!2B\! & \!\textbf{27.83}\!  &  \!{17.03}\! & \!{22.43}\! \\
    \!R1-SGG$^\star$\scriptsize{~\cite{chen2025compile}}\!  & \!7B\! & \!\textbf{28.77}\! &  \!{17.55}\! & \!{23.16}\! \\
    \textbf{Relation-R1$^\star$}\!(\textbf{Ours}) & \!3B\! & \!{25.87}\! &  \!\textbf{21.32}\! & \!\textbf{23.60}\! \\
    \hline
    \!ASMv2$^\dagger$\scriptsize{~\cite{wang2024all}}\! &  \!13B\!  & \!14.20\! & \!10.30\! & \!12.23\!\\ \!SpaceSGG$^\dagger$\scriptsize{~\cite{llava_spacesgg2025}} &  \!13B\!  & \!15.43\! & \!13.23\! & \!14.33\! \\
    \!\textbf{Relation-R1$^\dagger$}\! (\textbf{Ours}) & \!3B\! &  \!\textbf{22.33}\! &  \!\textbf{20.07}\! & \!\textbf{21.20}\! \\
    \hline
  \end{tabular}
  \caption{Comparison with various SGG models on PSG dataset~\cite{yang2022panoptic}. $\dagger$ denotes training/test with \textit{scene graph \textbf{caption} format} following ASMv2~\cite{wang2024all}. $\star$ indicates training/test in the \textit{standard scene graph format}$^{\ref{footnote:sgg_format}}$.}
    \label{tab:sgg_sota}
\end{table}

\noindent{\textbf{Evaluation Metrics.}}
 For \textit{binary relation detection},  we followed the standard SGG evaluation protocol~\cite{tang2020unbiased,wang2024all} 1) \textbf{Recall}: The ratio of correctly predicted triplets to the total number of ground-truth triplets, with both label accuracy and bounding box IoU >= 0.5 for subject and object. 2) \textbf{mean Recall} (\textbf{mRecall}): The average recall across all predicate classes. For \textit{\textit{N}-ary relation detection}, we adopted the same evaluation metrics as~\cite{pratt2020grounded}: 1) \textbf{verb}: the accuracy of verb prediction. 2) \textbf{value}: the noun accuracy for individual semantic role. 3) \textbf{value-all (val-all)}: Overall noun prediction accuracy across all semantic roles in an event. 4) \textbf{grounded-value (grnd)}: Bounding box accuracy for each semantic role, requiring an IoU >= 0.5 between predicted and ground-truth boxes. 5) \textbf{grounded-value-all (grnd-all)}: entire bounding box accuracy across all semantic roles in an event.

\noindent{\textbf{Implementation Details.}}
Refer to the Appendix.

\begin{table*}[!t]
    \centering
    \small
    \renewcommand{\arraystretch}{1.0}
     \begin{tabular}{|rl||c|c|c|c|c|c|}
    \thickhline
    \rowcolor{mygray}
    \multicolumn{2}{|c||}{Method} & Answer & Verb & Value & Value-all & Grnd & Grnd-all \\
    \hline
    ISL~\cite{pratt2020grounded} &{$_{ \text{ECCV'20}}$} & \multirow{6}{*}{\texttt{Close-ended}} & 39.36 & 30.09 & 18.62 & 22.73 & 7.72 \\
    JSL~\cite{pratt2020grounded} & {$_{ \text{ECCV'20}}$} & & 39.94 & 31.44 & 18.87 & 24.86 & 9.66 \\
    GSRTR~\cite{cho2021grounded} &{$_{ \text{BMVC'21}}$} & & 40.63 & 32.15 & 19.28 & 25.49 & 10.10 \\ 
    CoFormer~\cite{cho2022collaborative} &{$_{ \text{CVPR'22}}$} & & 44.66 & 35.98 & 22.22 & 29.05 & 12.21 \\
    SituFormer~\cite{wei2021rethinking} & {$_{ \text{AAAI'22}}$} & & 44.20 & 35.24 & 21.86 & 29.22 & 13.41 \\
    GSRFormer~\cite{cheng2022gsrformer} & {$_{ \text{ACM MM'22}}$} & & 46.53 & 37 .48 & 23.32 & 31.53 &  14.23 \\
    \hline
    OpenSU~\cite{liu2023opensu} & {$_{ \text{ICCVW'23}}$} & \multirow{2}{*}{\texttt{Open-ended}} &  50.10 & 41.20 & 26.56 & 34.27 & 15.70 \\
    \textbf{Relation-R1} (\textbf{Ours}) & & & \textbf{57.26}  &  \textbf{46.66} &  \textbf{30.92} &  \textbf{40.21} & \textbf{30.18} \\
    \hline
  \end{tabular}
    \caption{Performance(\%) comparison with various GSR models on the SWiG~\cite{pratt2020grounded} dataset.}
    \label{tab:gsr_sota}
\end{table*}

\begin{table*}[!t]
    \centering
    \small
    \renewcommand{\arraystretch}{1.0}
 \begin{tabular}{|l||c|cc|ccccc|} 
        \thickhline
        \rowcolor{mygray}
        & & \multicolumn{2}{c|}{Binary Relation} & 
        \multicolumn{5}{c|}{$N$-ary Relation}  \\
        \rowcolor{mygray}
        \multirow{-2}[0]{*}{Method} & \multirow{-2}[0]{*}{CoT} & Recall  & mRecall & Verb & Value & Value-all & Grnd & Grnd-all \\ 
        \hline
        SFT & - & 14.83 & 13.86 & 56.64 & 42.65~\scriptsize{(\textcolor{blue}{60.45})} & 22.11~\scriptsize{(\textcolor{blue}{25.21})} & 37.70~\scriptsize{(\textcolor{blue}{54.05})} & 16.35~\scriptsize{(\textcolor{blue}{19.02})} \\
        SFT + RL & \scriptsize{Template-based} & 20.24 & 17.31 & 58.38 & 47.75~\scriptsize{(\textcolor{blue}{66.74})}  & 32.00~\scriptsize{(\textcolor{blue}{35.67})}  & 41.27~\scriptsize{(\textcolor{blue}{55.88})} & 31.16~\scriptsize{(\textcolor{blue}{38.29})} \\
        SFT + RL & \scriptsize{MLLM-generated} & 20.66 & 20.30 & 53.00 & 42.27~\scriptsize{(\textcolor{blue}{63.86})} & 26.67~\scriptsize{(\textcolor{blue}{30.79})} & 35.69~\scriptsize{(\textcolor{blue}{52.43})} & 25.89~\scriptsize{(\textcolor{blue}{33.95})} \\
        SFT + RL & \scriptsize{Progressive} & 22.57 & 20.57 & 71.04 & 61.26~\scriptsize{(\textcolor{blue}{78.19})} & 44.98~\scriptsize{(\textcolor{blue}{49.73})} & 49.43~\scriptsize{(\textcolor{blue}{61.25})} & 36.09~\scriptsize{(\textcolor{blue}{42.35})}\\
        \hline
    \end{tabular}
    \caption{Performance comparison (\%) across various cognitive CoT strategies trained on single binary relation detection and $N$-ary relation detection, separately. \textcolor{blue}{\textbf{Blue}} indicates metrics without correct verb constraints.}
    \label{tab:cot}
\end{table*}

\begin{figure}[!t]
  \centering
\includegraphics[width=1.0\linewidth]{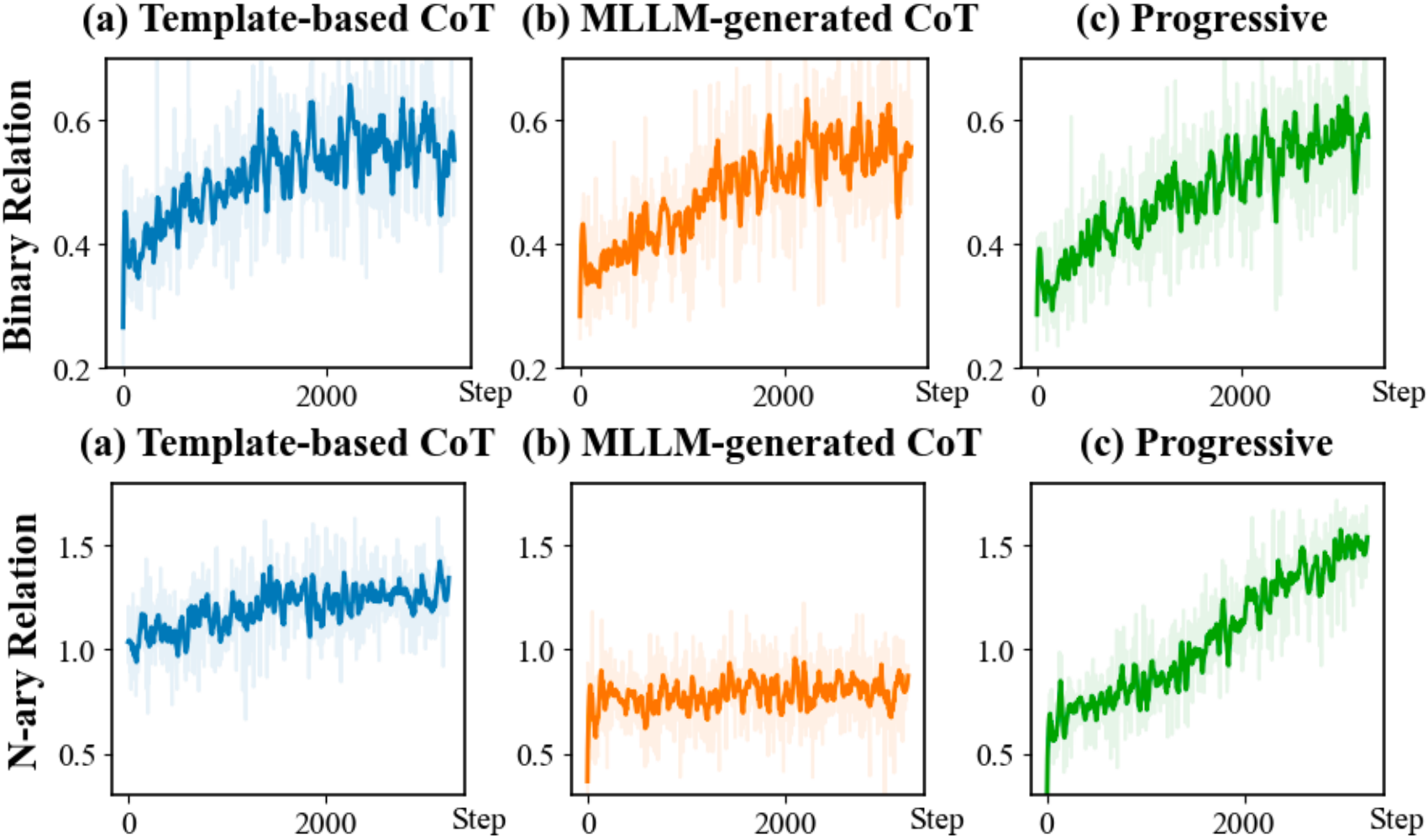}
  \caption{The statistics of binary and $N$-ary relation rewards.}
  \label{fig:reward}
\end{figure}

\subsection{Comparison with State-of-the-Art Methods}
\label{exp:quan}

We compared Relation-R1
for unified relation comprehension with the existing state-of-the-art binary relation detection (Table~\ref{tab:sgg_sota}) and \textit{N}-ary relation detection approaches in (Table~\ref{tab:gsr_sota}).

\noindent{\textbf{Binary Relation Detection.}} For the open-ended scene graph caption (\cf Table~\ref{tab:sgg_sota}\footnote{This Relation-R1 version is guided by the template-based CoTs.\label{footnote:sota_r1}}), Relation-R1 achieves state-of-the-art performance with the highest Recall of \textbf{22.33}\%, mRecall of \textbf{20.07}\% and overall Mean of \textbf{21.20}\%. This substantially surpasses prior open-ended methods such as LLaVA-SpaceSGG~\cite{llava_spacesgg2025} (15.43\% Recall, 13.23\% mRecall, 14.33\% Mean). Notably, even when evaluated on the standard scene graph format$^{\ref{footnote:sgg_format}}$, which presents a less challenging scenario than scene graph captioning due to the reduced need for contextual inference, Relation-R1 leads across all evaluation metrics. Furthermore, Relation-R1 achieves these results with a significantly smaller 3B parameter model, in contrast to the larger 13B parameter models employed by recent competitors ASMv2~\cite{wang2024all} and LLaVA-SpaceSGG, demonstrating superior parameter efficiency.

\begin{figure*}[!t]
    \centering
    \includegraphics[width=1.0\linewidth]
    {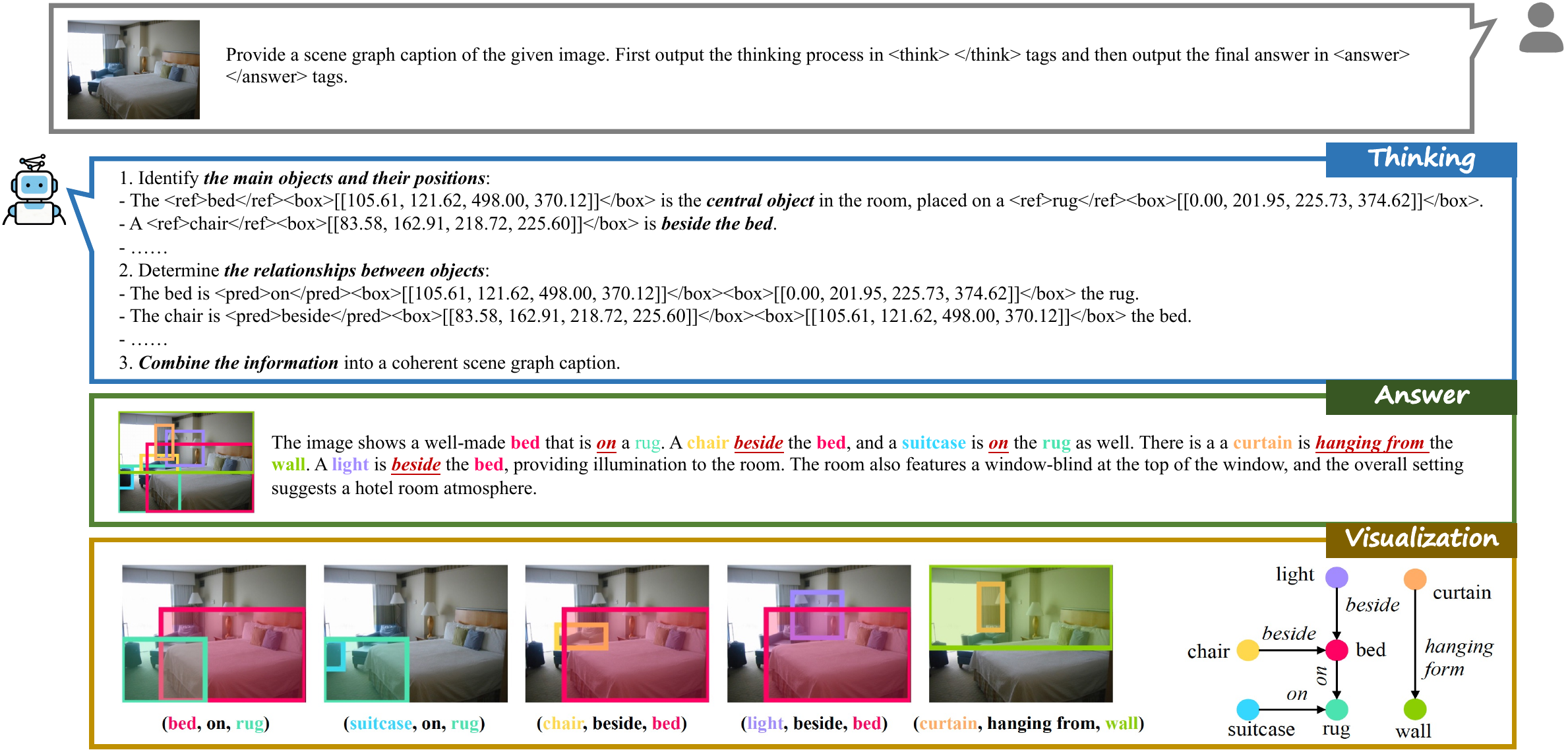}
    \caption{Qualitative results of binary relation detection.}
    \label{fig:sgg_vis}
\end{figure*}

\noindent{\textbf{\textit{N}-ary Relation Detection.}}
Within the open-ended grounded situation frame setting (\cf Table~\ref{tab:gsr_sota}$^{\ref{footnote:sota_r1}}$), Relation-R1 also establishes a new SOTA, achieving top performance across all metrics: Verb (\textbf{57.26}\%), Value (\textbf{46.66}\%), Value-all (\textbf{30.92}\%), Grnd (\textbf{40.21}\%), and Grnd-all (\textbf{30.18}\%). Notably, Relation-R1 significantly surpasses the previous best open-ended model, OpenSU~\cite{liu2023opensu}, exhibiting substantial gains, particularly with a remarkable \textbf{+14.48}\% absolute improvement on the challenging Grnd-all metric. This clearly demonstrates Relation-R1's superior capacity for comprehending and grounding complex \textit{N}-ary relationships in visual scenes.

\subsection{Diagnostic Experiment}
\label{sec:exp_diag}

To gain more insights and reduce the mutual influence between the two tasks during training, we conducted a set of comprehensive studies on the Relation-R1 trained with binary and \textit{N}-ary relation detection, respectively.

\noindent{\textbf{Cognitive Chain-of-Thoughts.}} Table~\ref{tab:cot} compares different cognitive CoT strategies for binary and \textit{N}-ary relation detection. As seen, introducing RL with various CoT strategies (SFT + RL) demonstrates clear benefits. The template-based CoT improves binary relation Recall to \textbf{20.24}\% and mRecall to \textbf{17.31}\%. For \textit{N}-ary relations, it enhances performance across the board compared to the constrained SFT baseline, for instance, achieving \textbf{47.75}\% on Value and \textbf{31.16}\% on Grnd-all. The MLLM-generated CoT slightly outperforms the template-based approach in binary relation detection, but underperforms in \textit{N}-ary relation detection. This is because \textit{N}-ary relation detection metrics are often limited by verb accuracy, the model may generate a verb different from the GT yet still semantically plausible. Notably, when the verb constraint is removed, MLLM-generated CoT yields significant gains, particularly in Grnd-all (+\textbf{14.93}\%). While template-based and MLLM-generated CoTs offer improvements over SFT, the progressive strategy stands out. This approach significantly outperforms all others across all metrics, achieving \textbf{22.57}\% Recall in binary relation detection and top scores (\eg, \textbf{71.04}\% Verb accuracy) in \textit{N}-ary relation detection. These results highlight the superior ability of the progressive CoT to guide complex relation detection.

\noindent{\textbf{Reward Analysis.}}
We visualized the reward variation curves for binary and \textit{N}-ary relation detection in Figure~\ref{fig:reward}, respectively. It can be seen that at the beginning stage, both tasks obtains decent rewards due to the foundation laid by SFT. With the gradual progress of GRPO, the rewards of both tasks show a gradually increasing trend. In binary relation detection, guided with progressive CoT yields a slightly enhanced overall reward trajectory, although the improvement is not significant. This might be since the gap between the CoT generated by MLLM (\eg, Qwen~\cite{Qwen2.5-VL}) and that produced by the template is not obvious. Conversely, \textit{N}-ary relation detection exhibits a significant surge in reward, primarily attributed to the model's emergent capability to generate synonymous grounded situation frames.

\noindent{\textbf{Completion Length Analysis.}}
\begin{figure}[!t]
  \centering
    \includegraphics[width=1.0\linewidth]{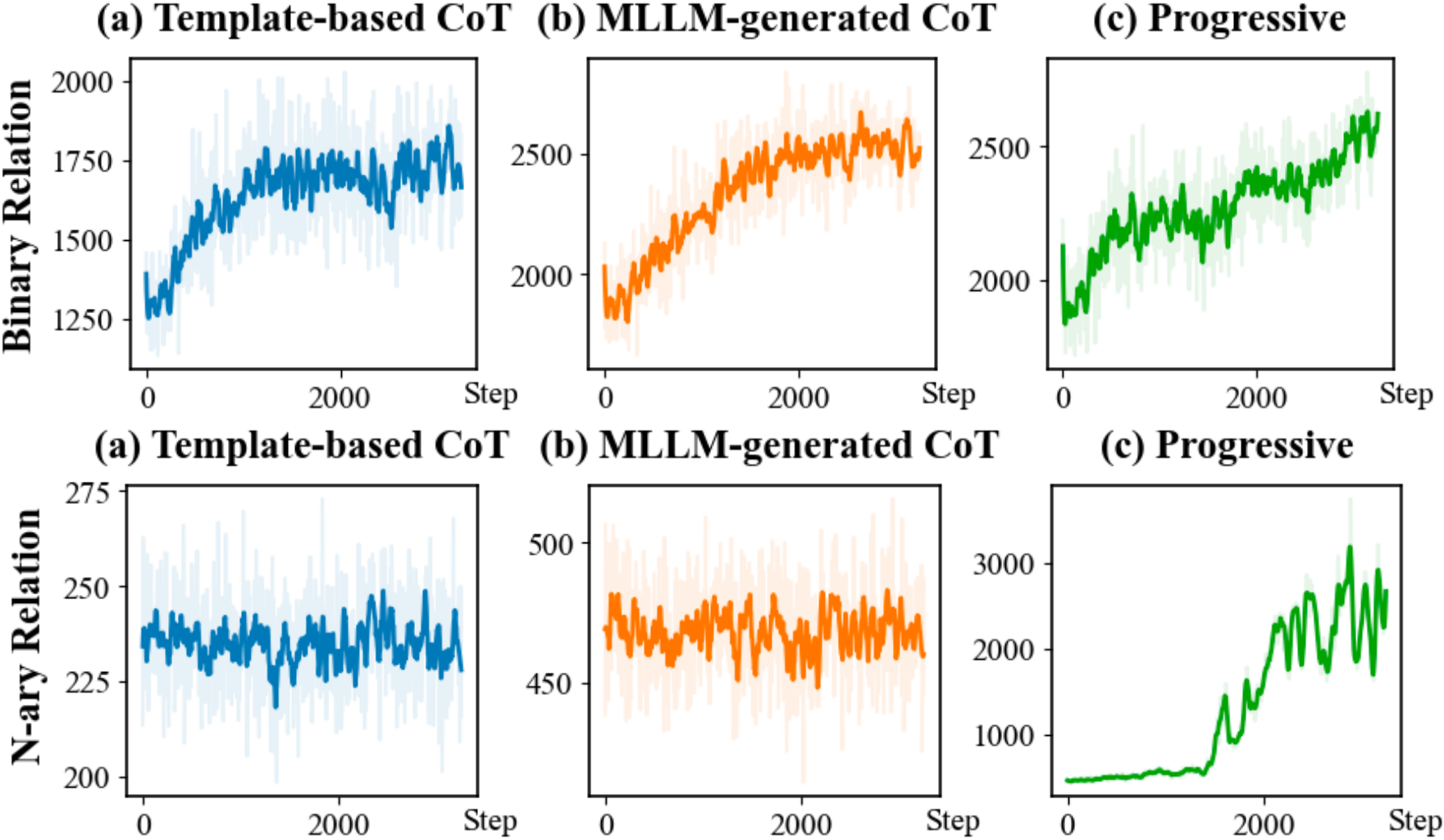}
  
  \caption{Completion lengths of binary and $N$-ary relation.}
  \label{fig:len}
\end{figure}
The statistic of completion length during GRPO training is in  Figure~\ref{fig:len}. Consistent with the reward trends, the completion lengths for binary relation detection exhibit a gradual increase as GRPO progresses, suggesting the generation of more detailed and descriptive outputs over time. Notably, the template-based CoT strategy yields significantly shorter completions compared to the more flexible MLLM-generated and progressive CoT strategies, since only the cognitive process is involved. The progressive CoT guidance, by enabling the model to explore synonymous relational expressions and incorporate more nuanced details, leads to an increase in the average completion length.

\noindent{\textbf{Qualitative Results.}}
Figure~\ref{fig:sgg_vis} shows a qualitative example of binary relation detection. As seen, the visualized thinking process reveals how Relation-R1 effectively decomposes the complex task into constituent cognitive steps: object classification, localization, and relation inference. For instance, after identifying the central \texttt{bed}, Relation-R1 pinpoints related objects \eg, \texttt{rug} and \texttt{chair}, and explicitly grounds the inferred spatial relations (\texttt{on}, \texttt{beside}) between them. Our approach demonstrates the successful decomposition of the scene into accurate, grounded object-relation pairs.

\section{Conclusion}
This paper reveals key limitations in MLLM visual relation understanding, particularly for \textit{N}-ary relations and language-induced bias. We propose Relation-R1, a unified two-stage framework: build a reasoning foundation and structured outputs via cognitive CoT-guided SFT, followed by GRPO, promoting visual-semantic grounding and robust generalization. Besides, progressively applied cognitive CoT guidance, transitioning from template-based to MLLM-generated reasoning, further enhances generalization, especially for synonymous \textit{N}-ary relations. Experiments on PSG and SWiG validate Relation-R1's effectiveness in both binary and \textit{N}-ary relation detection. Employing progressive CoT-guided SFT with R1-inspired refinement, Relation-R1 advances MLLMs towards more human-like relational reasoning.

\section{Acknowledgments}
This work was supported by the National Natural Science Foundation of China Young Scholar Fund (62402408), the Hong Kong SAR RGC General Research Fund under Grant (16208823), and the Hong Kong SAR RGC Early Career Scheme (26208924). This research was partially conducted by ACCESS – AI Chip Center for Emerging Smart Systems, supported by the InnoHK initiative of the Innovation and Technology Commission of the Hong Kong Special Administrative Region Government. This work was also supported by the Hong Kong SAR RGC General Research Fund under Grant (16208823) and the Hong Kong SAR RGC Early Career Scheme (26208924).

\bibliography{aaai2026}

@String(PAMI  = {IEEE Trans. Pattern Anal. Mach. Intell.})

@String(IJCV  = {Int. J. Comput. Vis.})

@String(CVPR  = {IEEE Conf. Comput. Vis. Pattern Recog.})

@String(ICCV  = {Int. Conf. Comput. Vis.})

@String(ECCV  = {Eur. Conf. Comput. Vis.})

@String(NeurIPS = {Adv. Neural Inform. Process. Syst.})

@String(ICML  = {Int. Conf. Mach. Learn.})

@String(ICLR  = {Int. Conf. Learn. Represent.})

@String(BMVC  = {Brit. Mach. Vis. Conf.})

@String(AAAI  = {AAAI})

@String(ICME  = {Int. Conf. Multimedia and Expo})

@String(ACMMM = {ACM Int. Conf. Multimedia})

@String(PAMI  = {IEEE TPAMI})

@String(IJCV  = {IJCV})

@String(CVPR  = {CVPR})

@String(ICCV  = {ICCV})

@String(ECCV  = {ECCV})

@String(WACV  = {WACV})

@String(NIPS = {NeurIPS})

@String(NeurIPS = {NeurIPS})

@String(ICML  = {ICML})

@String(ICLR  = {ICLR})

@String(BMVC  = {BMVC})

@String(ICME  = {ICME})

@String(ACMMM = {ACM MM})

@inproceedings{zellers2018neural,
  title={Neural motifs: Scene graph parsing with global context},
  author={Zellers, Rowan and Yatskar, Mark and Thomson, Sam and Choi, Yejin},
  booktitle=CVPR,
  pages={5831--5840},
  year={2018}
}

@inproceedings{li2022sgtr,
  title={Sgtr: End-to-end scene graph generation with transformer},
  author={Li, Rongjie and Zhang, Songyang and He, Xuming},
  booktitle=CVPR,
  pages={19486--19496},
  year={2022}
}

@inproceedings{xu2017scene,
  title={Scene graph generation by iterative message passing},
  author={Xu, Danfei and Zhu, Yuke and Choy, Christopher B and Fei-Fei, Li},
  booktitle=CVPR,
  pages={5410--5419},
  year={2017}
}

@inproceedings{Zhao_2023_ICCV,
    author    = {Zhao, Chengyang and Shen, Yikang and Chen, Zhenfang and Ding, Mingyu and Gan, Chuang},
    title     = {TextPSG: Panoptic Scene Graph Generation from Textual Descriptions},
    booktitle = ICCV,
    month     = {October},
    year      = {2023},
    pages     = {2839-2850}
}

@inproceedings{tang2019learning,
  title={Learning to compose dynamic tree structures for visual contexts},
  author={Tang, Kaihua and Zhang, Hanwang and Wu, Baoyuan and Luo, Wenhan and Liu, Wei},
  booktitle=CVPR,
  pages={6619--6628},
  year={2019}
}

@inproceedings{li2023zero,
  title={Zero-shot visual relation detection via composite visual cues from large language models},
  author={Li, Lin and Xiao, Jun and Chen, Guikun and Shao, Jian and Zhuang, Yueting and Chen, Long},
  booktitle=NeurIPS,
  year={2023}
}

@inproceedings{chen2020simple,
  title={A simple framework for contrastive learning of visual representations},
  author={Chen, Ting and Kornblith, Simon and Norouzi, Mohammad and Hinton, Geoffrey},
  booktitle=ICML,
  pages={1597--1607},
  year={2020},
}

@inproceedings{tang2020unbiased,
  title={Unbiased scene graph generation from biased training},
  author={Tang, Kaihua and Niu, Yulei and Huang, Jianqiang and Shi, Jiaxin and Zhang, Hanwang},
  booktitle=CVPR,
  pages={3716--3725},
  year={2020}
}

@inproceedings{li2022devil,
  title={The devil is in the labels: Noisy label correction for robust scene graph generation},
  author={Li, Lin and Chen, Long and Huang, Yifeng and Zhang, Zhimeng and Zhang, Songyang and Xiao, Jun},
  booktitle=CVPR,
  pages={18869--18878},
  year={2022}
}

@inproceedings{lin2020gps,
  title={Gps-net: Graph property sensing network for scene graph generation},
  author={Lin, Xin and Ding, Changxing and Zeng, Jinquan and Tao, Dacheng},
  booktitle=CVPR,
  pages={3746--3753},
  year={2020}
}

@article{yuan2024sr,
  author={Yuan, Zikang and Deng, Jie and Ming, Ruiye and Lang, Fengtian and Yang, Xin},
  journal={IEEE Robotics and Automation Letters}, 
  title={SR-LIVO: LiDAR-Inertial-Visual Odometry and Mapping With Sweep Reconstruction}, 
  year={2024},
  volume={9},
  number={6},
  pages={5110-5117}
}

@article{yuan2025voxel,
  author={Yuan, Zikang and Lang, Fengtian and Deng, Jie and Luo, Hongcheng and Yang, Xin},
  journal={IEEE Robotics and Automation Letters}, 
  title={Voxel-SVIO: Stereo Visual-Inertial Odometry based on Voxel Map}, 
  year={2025},
  volume={10},
  number={6},
  pages={6352-6359}
}

@inproceedings{chen2023addressing,
  title={Addressing Predicate Overlap in Scene Graph Generation with Semantic Granularity Controller},
  author={Chen, Guikun and Li, Lin and Luo, Yawei and Xiao, Jun},
  booktitle={ICME},
  year={2023}
}

@inproceedings{
lei2024seeing,
title={Seeing Beyond Classes: Zero-Shot Grounded Situation Recognition via Language Explainer},
author={Jiaming Lei and Lin Li and Chunping Wang and Jun Xiao and Long Chen},
booktitle=ACMMM,
year={2024},
}

@article{li2024nicest,
  title={Nicest: Noisy label correction and training for robust scene graph generation},
  author={Li, Lin and Xiao, Jun and Shi, Hanrong and Zhang, Hanwang and Yang, Yi and Liu, Wei and Chen, Long},
  journal=PAMI,
  year={2024},
  publisher={IEEE}
}

@inproceedings{li2023compositional,
  title={Compositional feature augmentation for unbiased scene graph generation},
  author={Li, Lin and Chen, Guikun and Xiao, Jun and Yang, Yi and Wang, Chunping and Chen, Long},
  booktitle=ICCV,
  pages={21685--21695},
  year={2023}
}

@inproceedings{wang2024all,
  title={The all-seeing project v2: Towards general relation comprehension of the open world},
  author={Wang, Weiyun and Ren, Yiming and Luo, Haowen and Li, Tiantong and Yan, Chenxiang and Chen, Zhe and Wang, Wenhai and Li, Qingyun and Lu, Lewei and Zhu, Xizhou and others},
  booktitle=ECCV,
  pages={471--490},
  year={2024},
  organization={Springer}
}

@inproceedings{llava_spacesgg2025,
  title={LLaVA-SpaceSGG: Visual Instruct Tuning for Open-vocabulary Scene Graph Generation with Enhanced Spatial Relations},
  author={Xu, Mingjie and Wu, Mengyang and Zhao, Yuzhi and Li, Jason Chun Lok and Ou, Weifeng},
  booktitle=WACV,
  year={2025}
}

@inproceedings{pratt2020grounded,
      title={Grounded situation recognition},
      author={Pratt, Sarah and Yatskar, Mark and Weihs, Luca and Farhadi, Ali and Kembhavi, Aniruddha},
      booktitle=ECCV,
      pages={314--332},
      year={2020}
}

@inproceedings {cho2021grounded,
      title={Grounded Situation Recognition with Transformers}, 
      author={Junhyeong Cho and Youngseok Yoon and Hyeonjun Lee and Suha Kwak},
      booktitle=BMVC,
      year={2021}

}

@inproceedings{yatskar2016situation,
  title={Situation recognition: Visual semantic role labeling for image understanding},
  author={Yatskar, Mark and Zettlemoyer, Luke and Farhadi, Ali},
  booktitle=CVPR,
  pages={5534--5542},
  year={2016}
}

@inproceedings{wei2021rethinking,
      title={Rethinking the Two-Stage Framework for Grounded Situation Recognition}, 
      author={Meng Wei and Long Chen and Wei Ji and Xiaoyu Yue and Tat-Seng Chua},
      booktitle=AAAI,
      pages={2651--2658},
      year={2022}
}

@inproceedings{cho2022collaborative,
      title={Collaborative Transformers for Grounded Situation Recognition}, 
      author={Junhyeong Cho and Youngseok Yoon and Suha Kwak},
      booktitle=CVPR,
      pages={19659--19668},
      year={2022}

}

@inproceedings{liu2023opensu,
  title={Open Scene Understanding: Grounded Situation Recognition Meets Segment Anything for Helping People with Visual Impairments},
  author={Liu, Ruiping and Zhang, Jiaming and Peng, Kunyu and Zheng, Junwei and Cao, Ke and Chen, Yufan and Yang, Kailun and Stiefelhagen, Rainer},
  booktitle={ICCVW},
  year={2023}
}

@article{DeepseekR1,
      title={DeepSeek-R1: Incentivizing Reasoning Capability in LLMs via Reinforcement Learning}, 
      author={DeepSeek-AI},
      journal={arXiv preprint arXiv:2501.12948},
      year={2025}
}

@article{schulman2017proximal,
  title={Proximal policy optimization algorithms},
  author={Schulman, John and Wolski, Filip and Dhariwal, Prafulla and Radford, Alec and Klimov, Oleg},
  journal={arXiv preprint arXiv:1707.06347},
  year={2017}
}

@inproceedings{yang2022panoptic,
  title={Panoptic scene graph generation},
  author={Yang, Jingkang and Ang, Yi Zhe and Guo, Zujin and Zhou, Kaiyang and Zhang, Wayne and Liu, Ziwei},
  booktitle=ECCV,
  pages={178--196},
  year={2022},
  organization={Springer}
}

@article{liu2025visual,
  title={Visual-rft: Visual reinforcement fine-tuning},
  author={Liu, Ziyu and Sun, Zeyi and Zang, Yuhang and Dong, Xiaoyi and Cao, Yuhang and Duan, Haodong and Lin, Dahua and Wang, Jiaqi},
  journal={arXiv preprint arXiv:2503.01785},
  year={2025}
}

@article{yang2025r1onevision,
  title={R1-Onevision: Advancing Generalized Multimodal Reasoning through Cross-Modal Formalization},
  author={Yi Yang and Xiaoxuan He and Hongkun Pan and Xiyan Jiang and Yan Deng and Xingtao Yang and Haoyu Lu and Dacheng Yin and Fengyun Rao and Minfeng Zhu and Bo Zhang and Wei Chen},
  journal={arXiv preprint arXiv:2503.10615},
  year={2025},
}

@article{du2025virgo,
  title={Virgo: A Preliminary Exploration on Reproducing o1-like MLLM},
  author={Du, Yifan and Liu, Zikang and Li, Yifan and Zhao, Wayne Xin and Huo, Yuqi and Wang, Bingning and Chen, Weipeng and Liu, Zheng and Wang, Zhongyuan and Wen, Ji-Rong},
  journal={arXiv preprint arXiv:2501.01904},
  year={2025}
}

@article{huang2025vision,
  title={Vision-r1: Incentivizing reasoning capability in multimodal large language models},
  author={Huang, Wenxuan and Jia, Bohan and Zhai, Zijie and Cao, Shaosheng and Ye, Zheyu and Zhao, Fei and Hu, Yao and Lin, Shaohui},
  journal={arXiv preprint arXiv:2503.06749},
  year={2025}
}

@article{meng2025mm,
  title={MM-Eureka: Exploring Visual Aha Moment with Rule-based Large-scale Reinforcement Learning},
  author={Meng, Fanqing and Du, Lingxiao and Liu, Zongkai and Zhou, Zhixiang and Lu, Quanfeng and Fu, Daocheng and Shi, Botian and Wang, Wenhai and He, Junjun and Zhang, Kaipeng and others},
  journal={arXiv preprint arXiv:2503.07365},
  year={2025}
}

@article{zhao2025r1,
  title={R1-omni: Explainable omni-multimodal emotion recognition with reinforcement learning},
  author={Zhao, Jiaxing and Wei, Xihan and Bo, Liefeng},
  journal={arXiv e-prints},
  pages={arXiv--2503},
  year={2025}
}

@article{liu2025seg,
  title={Seg-zero: Reasoning-chain guided segmentation via cognitive reinforcement},
  author={Liu, Yuqi and Peng, Bohao and Zhong, Zhisheng and Yue, Zihao and Lu, Fanbin and Yu, Bei and Jia, Jiaya},
  journal={arXiv preprint arXiv:2503.06520},
  year={2025}
}

@article{jaech2024openai,
  title={Openai o1 system card},
  author={Jaech, Aaron and Kalai, Adam and Lerer, Adam and Richardson, Adam and El-Kishky, Ahmed and Low, Aiden and Helyar, Alec and Madry, Aleksander and Beutel, Alex and Carney, Alex and others},
  journal={arXiv preprint arXiv:2412.16720},
  year={2024}
}

@inproceedings{cheng2022gsrformer,
  title={Gsrformer: Grounded situation recognition transformer with alternate semantic attention refinement},
  author={Cheng, Zhi-Qi and Dai, Qi and Li, Siyao and Mitamura, Teruko and Hauptmann, Alexander},
  booktitle={ACM MM},
  pages={3272--3281},
  year={2022}
}

@article{chen2024scene,
  title={Scene graph generation with role-playing large language models},
  author={Chen, Guikun and Li, Jin and Wang, Wenguan},
  journal=NIPS,
  year={2024}
}

@inproceedings{li2024pixels,
  title={From Pixels to Graphs: Open-Vocabulary Scene Graph Generation with Vision-Language Models},
  author={Li, Rongjie and Zhang, Songyang and Lin, Dahua and Chen, Kai and He, Xuming},
  booktitle={CVPR},
  pages={28076--28086},
  year={2024}
}

@inproceedings{chen2024expanding,
  title={Expanding Scene Graph Boundaries: Fully Open-vocabulary Scene Graph Generation via Visual-Concept Alignment and Retention},
  author={Chen, Zuyao and Wu, Jinlin and Lei, Zhen and Zhang, Zhaoxiang and Chen, Changwen},
  booktitle=ECCV,
  year={2024}
}

@article{chu2025sft,
  title={Sft memorizes, rl generalizes: A comparative study of foundation model post-training},
  author={Chu, Tianzhe and Zhai, Yuexiang and Yang, Jihan and Tong, Shengbang and Xie, Saining and Schuurmans, Dale and Le, Quoc V and Levine, Sergey and Ma, Yi},
  journal={arXiv preprint arXiv:2501.17161},
  year={2025}
}

@article{Qwen2.5-VL,
  title={Qwen2.5-VL Technical Report},
  author={Bai, Shuai and Chen, Keqin and Liu, Xuejing and Wang, Jialin and Ge, Wenbin and Song, Sibo and Dang, Kai and Wang, Peng and Wang, Shijie and Tang, Jun and Zhong, Humen and Zhu, Yuanzhi and Yang, Mingkun and Li, Zhaohai and Wan, Jianqiang and Wang, Pengfei and Ding, Wei and Fu, Zheren and Xu, Yiheng and Ye, Jiabo and Zhang, Xi and Xie, Tianbao and Cheng, Zesen and Zhang, Hang and Yang, Zhibo and Xu, Haiyang and Lin, Junyang},
  journal={arXiv preprint arXiv:2502.13923},
  year={2025}
}

@inproceedings{carion2020end,
    title={End-to-end object detection with transformers},
    author={Carion, Nicolas and Massa, Francisco and Synnaeve, Gabriel and Usunier, Nicolas and Kirillov, Alexander and Zagoruyko, Sergey},
    booktitle=ECCV,
    year={2020}
}

@inproceedings{zhang2024llava,
  title={Llava-grounding: Grounded visual chat with large multimodal models},
  author={Zhang, Hao and Li, Hongyang and Li, Feng and Ren, Tianhe and Zou, Xueyan and Liu, Shilong and Huang, Shijia and Gao, Jianfeng and Leizhang and Li, Chunyuan and others},
  booktitle=ECCV,
  pages={19--35},
  year={2024},
  organization={Springer}
}

@inproceedings{youferret,
  title={Ferret: Refer and Ground Anything Anywhere at Any Granularity},
  author={You, Haoxuan and Zhang, Haotian and Gan, Zhe and Du, Xianzhi and Zhang, Bowen and Wang, Zirui and Cao, Liangliang and Chang, Shih-Fu and Yang, Yinfei},
  booktitle=ICLR,
  year={2023}
}

@article{liu2023visual,
  title={Visual instruction tuning},
  author={Liu, Haotian and Li, Chunyuan and Wu, Qingyang and Lee, Yong Jae},
  journal=NIPS,
  volume={36},
  pages={34892--34916},
  year={2023}
}

@inproceedings{zhu2024minigpt,
  title={MiniGPT-4: Enhancing Vision-Language Understanding with Advanced Large Language Models},
  author={Zhu, Deyao and Chen, Jun and Shen, Xiaoqian and Li, Xiang and Elhoseiny, Mohamed},
  booktitle=ICLR,
  year={2024}
}

@inproceedings{lai2024lisa,
  title={Lisa: Reasoning segmentation via large language model},
  author={Lai, Xin and Tian, Zhuotao and Chen, Yukang and Li, Yanwei and Yuan, Yuhui and Liu, Shu and Jia, Jiaya},
  booktitle=CVPR,
  pages={9579--9589},
  year={2024}
}

@inproceedings{rasheed2024glamm,
  title={Glamm: Pixel grounding large multimodal model},
  author={Rasheed, Hanoona and Maaz, Muhammad and Shaji, Sahal and Shaker, Abdelrahman and Khan, Salman and Cholakkal, Hisham and Anwer, Rao M and Xing, Eric and Yang, Ming-Hsuan and Khan, Fahad S},
  booktitle=CVPR,
  pages={13009--13018},
  year={2024}
}

@inproceedings{peng2024kosmos,
  title={Kosmos-2: Grounding Multimodal Large Language Models to the World},
  author={Peng, Zhiliang and Wang, Wenhui and Dong, Li and Hao, Yaru and Huang, Shaohan and Ma, Shuming and Wei, Furu},
  booktitle=ICLR,
  year={2024}
}

@article{chen2025compile,
  title={Compile Scene Graphs with Reinforcement Learning},
  author={Chen, Zuyao and Wu, Jinlin and Lei, Zhen and Pollefeys, Marc and Chen, Chang Wen},
  journal={arXiv preprint arXiv:2504.13617},
  year={2025}
}

@inproceedings{shen2025tarpro,
  title={Tarpro: Targeted protection against malicious image editing},
  author={Shen, Kaixin and Quan, Ruijie and Miao, Jiaxu and Xiao, Jun and Yang, Yi},
  booktitle= {AAAI},
  year={2026}
}

@article{shao2024knowledge,
  title={Knowledge-guided causal intervention for weakly-supervised object localization},
  author={Shao, Feifei and Luo, Yawei and Gao, Fei and Yang, Yi and Xiao, Jun},
  journal={IEEE Transactions on Knowledge and Data Engineering},
  volume={36},
  number={11},
  pages={6477--6489},
  year={2024},
  publisher={IEEE}
}

@article{li2025inter,
  title={Interaction-Centric Knowledge Infusion and Transfer for Open Vocabulary Scene Graph Generation},
  author={Li, Lin and Zhang, Chuhan and Zhang, Dong and Sun, Chong and Li, Chen and Chen, Long},
  journal={Advances in Neural Information Processing Systems},
  year={2025}
}

@article{shi2025easy,
  title={From easy to hard: Learning curricular shape-aware features for robust panoptic scene graph generation},
  author={Shi, Hanrong and Li, Lin and Xiao, Jun and Zhuang, Yueting and Chen, Long},
  journal=IJCV,
  volume={133},
  number={1},
  pages={489--508},
  year={2025},
  publisher={Springer}
}
\clearpage
\appendix
\setcounter{secnumdepth}{3}
\renewcommand{\thetable}{S\arabic{table}}
\renewcommand{\thefigure}{S\arabic{figure}}
\setcounter{table}{0}
\setcounter{figure}{0}

\section*{Summary of the Appendix}
To enhance the understanding of the main paper, supplementary material containing further details are provided, organized as follows:
\begin{itemize} 
    \item \S\ref{sec:supp_detail} elaborates on the implementation details of Relation-R1. 
    \item \S\ref{supp_sec:cot_gen} introduces the cognitive CoT generation prompt. 
    \item \S\ref{supp_sec:sgg_format} details the scene graph format utilized in this work.
    \item \S\ref{supp_sec:exp} provides additional experimental results. 
    \item \S\ref{supp_sec:vis} shows further qualitative results. 
    \item \S\ref{supp_sec:limits} discusses limitations and broader impacts of Relation-R1.
\end{itemize}
\section{Implementation Details}
\label{sec:supp_detail}
We utilized Qwen2.5-VL-3B~\cite{Qwen2.5-VL} as our base model. Relation-R1 was trained on an 8xA100 80GB GPU server. During the SFT phase, we employed a total batch size of 32, with 4 samples processed per training step. For the GRPO component, we used a total batch size of 48. The learning rates were set to 2e-5 for SFT and 1e-6 for GRPO, respectively. Qwen2.5-VL-72B~\cite{Qwen2.5-VL} served as the CoT generator for producing MLLM-generated CoTs. The weighting hyperparameters $\alpha$ and $\beta$ were all set to 0.5 in the reward function. Progressive training began with supervised fine-tuning on binary and \textit{N}-ary relation detection tasks for 2 epochs using template-based CoT, followed by fine-tuning on 4k MLLM-generated CoT samples. After that, GRPO RL training was conducted: 2.4k steps on the \textit{N}-ary relation detection task, 3.6k steps on the binary relation detection task, and finally 2k steps on both tasks.
\section{Cognitive CoT Generation Prompt}
\label{supp_sec:cot_gen}
\begin{figure*}[!htpb]
\small
  \begin{tcolorbox}[left=0mm, right=5mm]
    \footnotesize
    \begin{tabular}{p{1.0\linewidth}} 
        \VarSty{ {\bf Question:}} 
Scene graph caption is a textual description of an image that incorporates references for objects and their relationships:

1. Use <ref>\texttt{object class}</ref><box>[[x1,y1,x2,y2],...,[x1,y1,x2,y2]]</box> to denote all subject and object regions of a relation triplet in the caption. 

2. Use <pred>\texttt{predicate class}</pred><box>subject coordinates list</box><box>object coordinates list</box> to denote all relation triplet regions.

3. Each sentence must have at least one relation triplet region. 

Given the image, a reasonable scene graph caption is:

\{\textbf{\texttt{ground-truth caption}}\}

To make you generate a reasonable scene graph caption, please give the step-by-step reasoning process in <think> </think> tags. \\

    \end{tabular}
\end{tcolorbox}
  \vspace{-1.5em}
  \captionsetup{font=small} 
  \caption{Scene graph caption CoT generation prompt.}
\label{tab:qw_sgg} 
\end{figure*}

\begin{figure*}[!htpb]
\small
  \begin{tcolorbox}[left=0mm, right=5mm]
    \footnotesize
    \begin{tabular}{p{1.0\linewidth}} 
        \VarSty{ {\bf Question:}} Grounded situation frame is a multimodal description combining textual semantics with visual localization information, where key situational elements (agent, action, object, place) are associated with both semantic role labels and corresponding bounding box coordinates in the image. It follows the pattern: <role>\texttt{entity class}</role><box>[x1,y1,x2,y2]</box>.

Given the image, a reasonable grounded situation frame is:

\{\textbf{\texttt{ground-truth frame}}\}

To make you generate a reasonable grounded situation frame, please give the step-by-step reasoning process in <think> </think> tags. \\
    \end{tabular}
\end{tcolorbox}
  \vspace{-1.5em}
  \captionsetup{font=small} 
  \caption{Grounded situation frame CoT generation prompt.}
\label{tab:qw_gsr} 
\end{figure*}
As mentioned above, we employ two distinct prompts (Figure~\ref{tab:qw_sgg} and Figure~\ref{tab:qw_gsr}) to guide cognitive CoT generation for binary and \textit{N}-ary relation detection, respectively. Each prompt strategically comprises: 1) a concise task definition outlining the detection objective; 2) an illustrative in-context example demonstrating the desired reasoning steps and output format; and 3) a specific instruction prompting the model to generate CoT of the current image.
\section{Scene Graph Format}
\label{supp_sec:sgg_format}
\begin{figure*}[!htpb]
\small
  \begin{tcolorbox}[left=0mm, right=5mm]
    \footnotesize
    \begin{tabular}{p{1.0\linewidth}} 
        \VarSty{ {\bf Question:}} Generate a scene graph for the provided image as a structured list containing [subject entity, [x1,y1,x2,y2] bounding box coordinates of subject, object entity, [x1,y1,x2,y2] bounding box coordinates of object, relationships between subject and object]. First output the thinking process in <think> </think> tags and then output the final answer in <answer> </answer> tags. \\
    \end{tabular}
\end{tcolorbox}
  \vspace{-1em}
  \captionsetup{font=small} 
  \caption{Scene graph format prompt used for training.}
\label{tab:qw_sgg_format} 
\end{figure*}
The scene graph format prompt employed for GRPO training is displayed in Figure~\ref{tab:qw_sgg_format}. Furthermore, to enable the model's generation of the think process, we utilize template-based CoT for supervised fine-tuning.

\section{More Experimental Results}
\label{supp_sec:exp}
\begin{table*}[!t]
    \centering
    \small
    \captionsetup{font=small}
    \caption{Few-Shot GSR performance comparison. 
    }
    \vspace{-0.5em}
    \label{tab:gsr_fewshot}
    \scalebox{0.95}{
    \setlength\tabcolsep{6pt}
     \begin{tabular}{|rl||c|c|c|c|c|c|}
    \thickhline
    \rowcolor{mygray}
    \multicolumn{2}{|c||}{Method} & Setting & Verb & Value & Value-all & Grnd & Grnd-all \\
    \hline
    \hline
    GSRTR~\cite{cho2021grounded} &{$_{ \text{BMVC'21}}$} & \multirow{4}{*}{\texttt{One-Shot}} &  0.35 & 0.10 & 0.01 & 0.08 & 0.00  \\ 
    CoFormer~\cite{cho2022collaborative} &{$_{ \text{CVPR'22}}$} & & 0.63 & 0.23 & 0.02 & 0.14 & 0.01 \\
    \textbf{Relation-R1} (\textbf{Ours}) & &  & 31.67 & 15.37 & 4.15 & 14.01 & 3.41 \\
    $w$/$o$ RL & &  & 27.31 & 10.21 & 1.43 & 9.35 & 1.22 \\

    \hline
    \hline
    GSRTR~\cite{cho2021grounded} &{$_{ \text{BMVC'21}}$} & \multirow{4}{*}{\texttt{4-Shot}} &  6.95 & 4.32 & 1.83 & 2.79 & 0.65  \\ CoFormer~\cite{cho2022collaborative} &{$_{ \text{CVPR'22}}$} & & 12.39	& 8.27	& 3.93 & 5.28 & 1.25  \\
    \textbf{Relation-R1} (\textbf{Ours}) & &  & 38.02 & 21.87 & 7.39 & 19.64 & 5.92\\
    $w$/$o$ RL & &  & 34.88 & 17.14 & 3.94 & 15.63 & 3.38 \\

    \hline
  \end{tabular}
  }
\end{table*}
\noindent\underline{\textbf{Few-Shot \textit{N}-ary Relation Detection.}}
We also evaluated Relation-R1's few-shot generalization capability (Table \ref{tab:gsr_fewshot}). In the one-shot setting, Relation-R1 achieved dramatic improvements over prior work, reaching \textbf{31.67}\% (Verb) and \textbf{3.41}\% (Grnd-all) compared to CoFormer's~\cite{cho2022collaborative} 0.63\% and 0.01\%, respectively. This advantage is maintained in the 4-shot setting, where Relation-R1 reached \textbf{38.02}\% (Verb) and \textbf{5.92}\% (Grnd-all), significantly surpassing previous methods. These results clearly demonstrate Relation-R1's superior ability to generalize and perform robust \textit{N}-ary relation detection, setting a new SOTA for few-shot GSR.

\noindent\underline{\textbf{Joint Binary \& \textit{N}-ary Relation Detection.}}
    \begin{table*}[!t]
        \centering
        \small
        \captionsetup{font=small}
        \caption{Performance comparison (\%) across various cognitive CoT strategies trained on \textit{joint} binary relation detection and $N$-ary relation detection, separately. \textcolor{blue}{\textbf{Blue}} indicates metrics without correct verb constraints.}
        \vspace{-0.5em}
        \label{tab:merge_cot}
        \scalebox{0.95}{
        \setlength\tabcolsep{6pt}
     \begin{tabular}{|l||c|cc|ccccc|} 
            \thickhline
            \rowcolor{mygray}
            & & \multicolumn{2}{c|}{Binary Relation} & 
            \multicolumn{5}{c|}{$N$-ary Relation}  \\
            \rowcolor{mygray}
            \multirow{-2}[0]{*}{Method} & \multirow{-2}[0]{*}{CoT} & Recall  & mRecall & Verb & Value & Value-all & Grnd & Grnd-all \\ 
            \hline
            \hline
            SFT & - & 14.89 & 12.66 & 56.23 & 45.81~\scriptsize{(\textcolor{blue}{66.00})} & 30.31~\scriptsize{(\textcolor{blue}{34.06})} & 38.59~\scriptsize{(\textcolor{blue}{54.11})} & 28.77~\scriptsize{(\textcolor{blue}{36.18})} \\
            SFT + RL & \scriptsize{Template-based} & 22.33 & 20.07 & 57.26 & 46.66~\scriptsize{(\textcolor{blue}{66.14})}  & 30.92~\scriptsize{(\textcolor{blue}{34.75})}  & 40.21~\scriptsize{(\textcolor{blue}{55.50})} & 30.18~\scriptsize{(\textcolor{blue}{37.64})} \\
            SFT + RL & \scriptsize{MLLM-generated} & 19.59 & 18.30 & 53.00 & 42.27~\scriptsize{(\textcolor{blue}{65.11})} & 26.67~\scriptsize{(\textcolor{blue}{34.65})} & 35.69~\scriptsize{(\textcolor{blue}{55.36})} & 25.89~\scriptsize{(\textcolor{blue}{37.61})} \\
            SFT + RL & \scriptsize{Progressive} & \textbf{27.19} & \textbf{26.12} & \textbf{73.25} & \textbf{66.18}~\scriptsize{(\textcolor{blue}{82.54})} & \textbf{52.90}~\scriptsize{(\textcolor{blue}{57.83})} & \textbf{55.20}~\scriptsize{(\textcolor{blue}{67.13})} & \textbf{42.36}~\scriptsize{(\textcolor{blue}{48.81})}\\
            \hline
        \end{tabular}
        }
    \end{table*}
Beyond solely training Relation-R1 on binary or \textit{N}-ary relation detection tasks, we also investigated the joint training of both tasks within the same Relation-R1 model. The results of this joint training are presented in Table~\ref{tab:merge_cot}\footnote{The Grnd-all metric excludes ground-truth bounding boxes represented as [-1, -1, -1, -1] for calculation.\label{footnote:grnd-all}}. From the analysis of these results, three primary conclusions can be drawn: 1) When utilizing both the SFT and RL strategies, our Relation-R1 consistently achieves higher performance across nearly all metrics for both binary and \textit{N}-ary relation detection tasks. Similar to single-task training, the MLLM-generated CoT strategy underperforms in the \textit{N}-ary task. This is likely due to the inherent generality of the CoT generated by Qwen~\cite{Qwen2.5-VL}, which may not precisely align with the ground-truth annotations, consequently leading to lower Verb accuracy. 2) Our proposed progressive CoT-guided paradigm consistently achieves the highest performance across all evaluated metrics for both tasks in the joint learning setting. 3) The joint progressive CoT-guided training across both binary and \textit{N}-ary tasks fosters a synergistic mutual promotion effect. This is evidenced by the notable gains observed: for instance, a Recall of \textbf{27.19}\% in Table~\ref{tab:merge_cot} \vs 22.57\% in Table~\ref{tab:cot} (for binary relation detection), and a Verb accuracy of \textbf{73.25}\% in Table~\ref{tab:merge_cot} \vs 71.04\% in Table~\ref{tab:cot} (for \textit{N}-ary relation detection).

\section{Qualitative Results} 
\label{supp_sec:vis}

\begin{figure*}[t]
  \centering
\includegraphics[width=\linewidth]{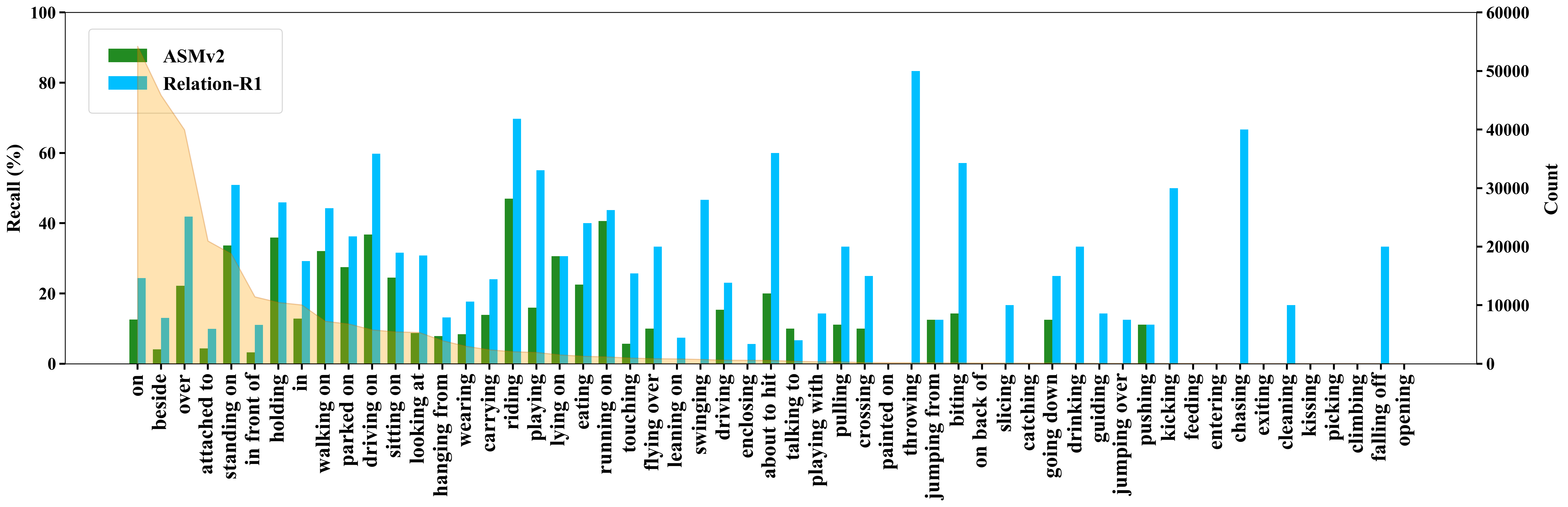}
\captionsetup{font=small}
\vspace{-1.0em}
\caption{Performance(\%) comparison between ASMv2 and Relation-R1 over all predicates on test set of PSG~\cite{yang2022panoptic} dataset. The \textcolor{myorange}{orange} area denotes the predicate distribution of the training set.} 
\label{fig:sgg_distribution}
\end{figure*}
\begin{figure*}[!t]
    \centering
    \includegraphics[width=0.90\linewidth]
    {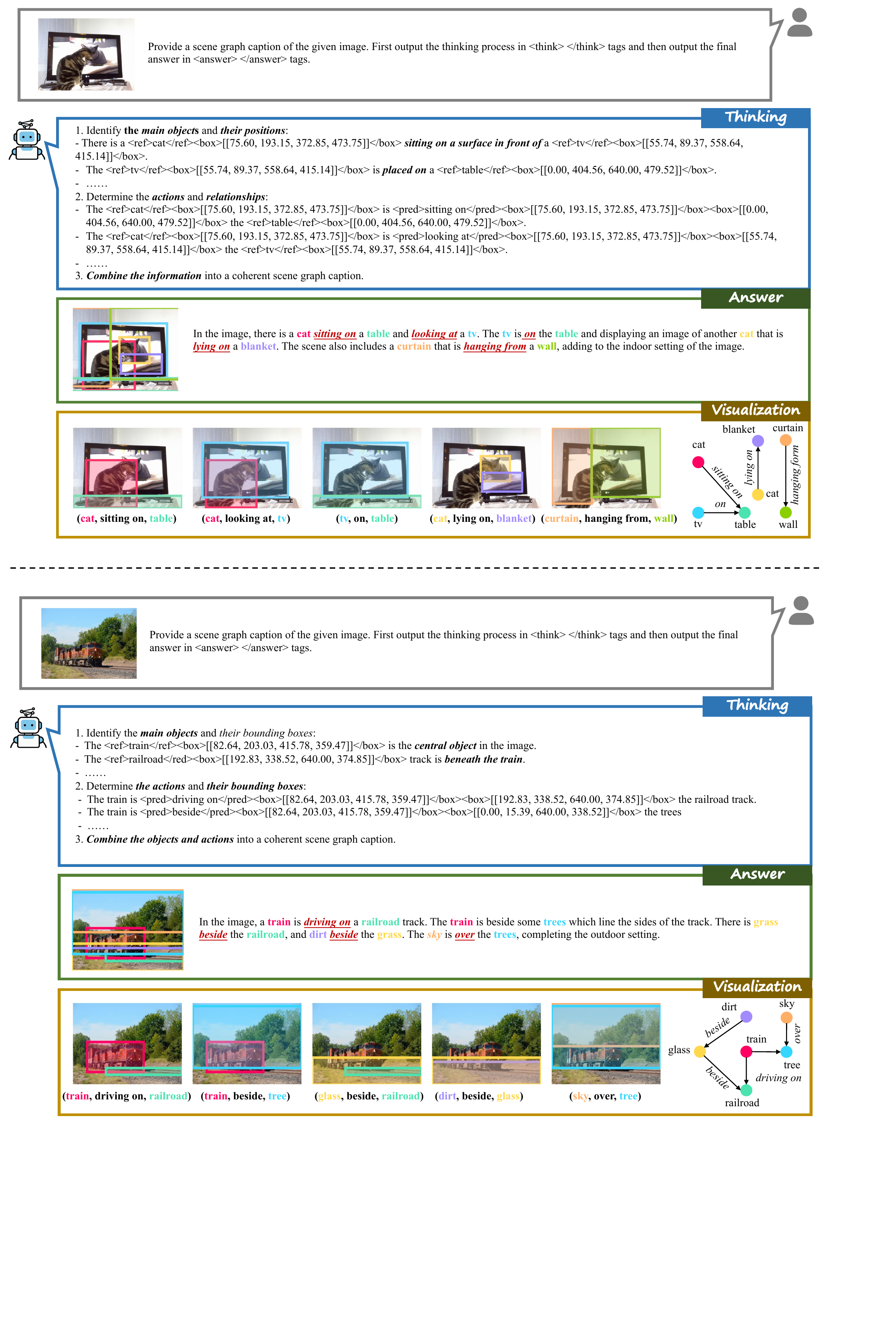}
    \vspace{-0.5em}
    \captionsetup{font=small}
    \caption{Qualitative results of binary relation detection.}
    \vspace{-1em}
    \label{fig:sgg_supp_vis}
\end{figure*}
\begin{figure*}[!htpb]
    \centering
    \includegraphics[width=0.85\linewidth]{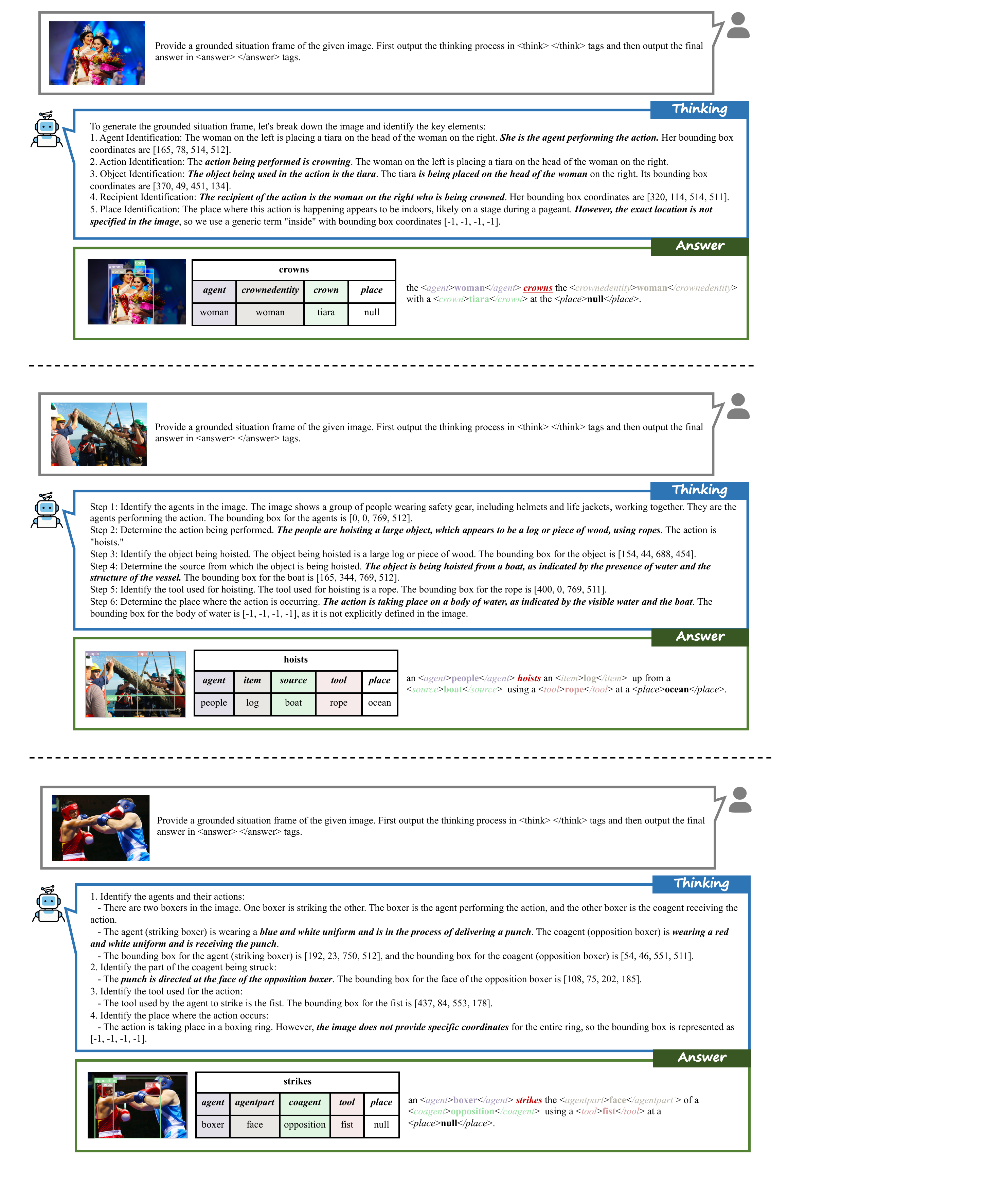}
    \vspace{-0.5em}
    \captionsetup{font=small}
    \caption{Qualitative results of \textit{N}-ary relation detection.}
    \label{fig:gsr_vis}
\end{figure*}
\begin{figure*}[!htpb]
    \centering
    \includegraphics[width=0.85\linewidth]{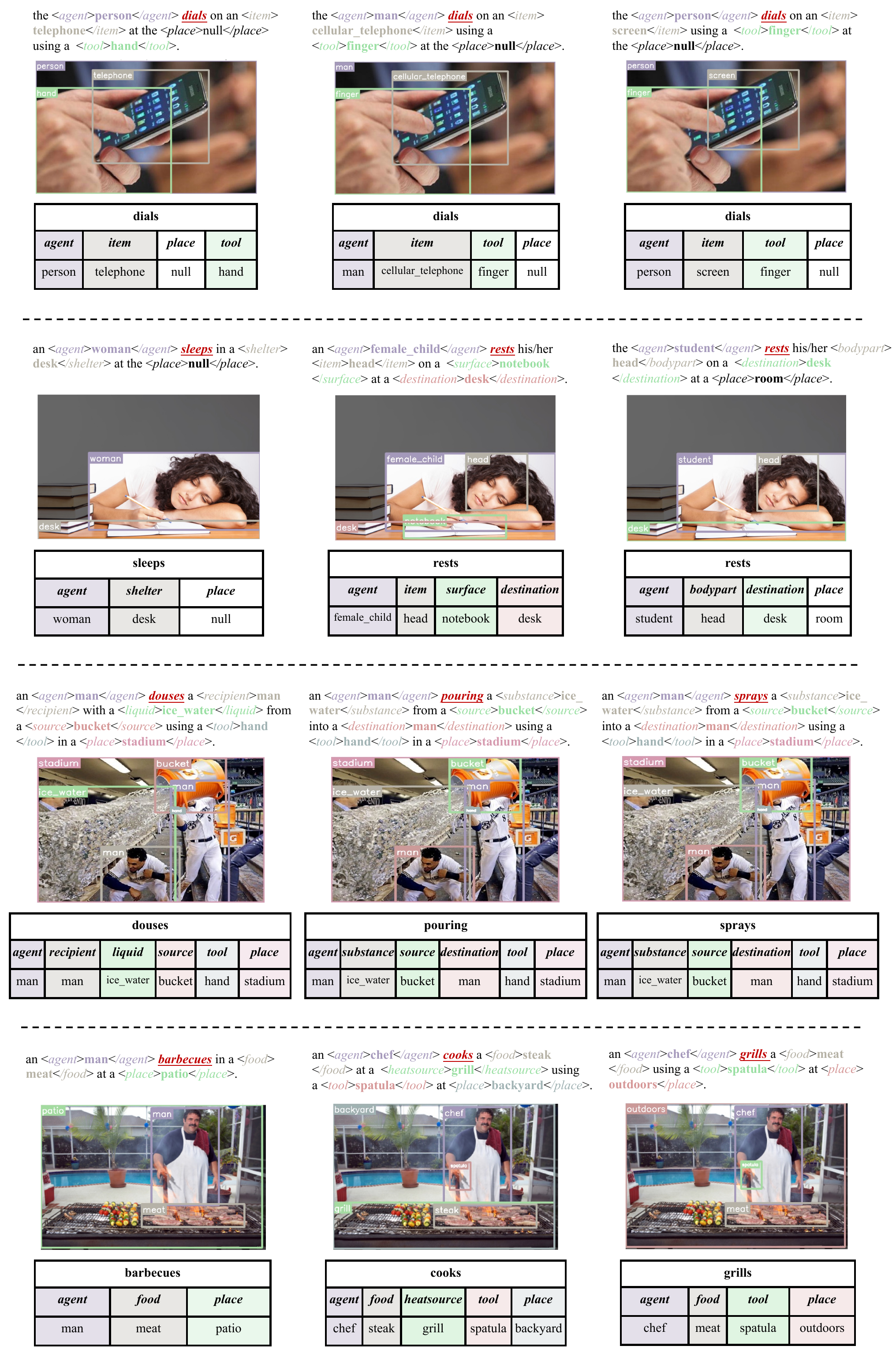}
    \vspace{-0.5em}
    \captionsetup{font=small}
    \caption{Qualitative results of \textit{synonymous} in \textit{N}-ary relation detection.}
    \label{fig:syn}
\end{figure*}
\noindent\underline{\textbf{Comparison over All Predicates.}}
We visualized predicate-level comparison against ASMv2~\cite{wang2024all} and the predicate distribution in Figure~\ref{fig:sgg_distribution}. As seen, Relation-R1 outperforms ASMv2 by a clear margin on both prevalent and rare predicates. These results clearly validated the effectiveness and efficiency of our approach for the SGG task in the open-ended setting.

\noindent\underline{\textbf{Binary and \textit{N}-ary Relation Detection Results.}} 
We visualized some examples of binary and \textit{N}-ary relation detection in Figure~\ref{fig:sgg_supp_vis} and Figure~\ref{fig:gsr_vis}, respectively. As seen, for \textbf{binary relation detection} (\cf Figure~\ref{fig:sgg_supp_vis}), the visualized thinking process reveals how Relation-R1 effectively decomposes the complex task into constituent cognitive steps: object classification, localization, and relation inference. For instance, Relation-R1 can first focus on the \textit{main objects} (\eg, ``\texttt{cat}'' and ``\texttt{train}'') and then infer their positions and relationships.
Similarly, for \textbf{\textit{N}-ary relation detection} (\cf Figure~\ref{fig:gsr_vis}), Relation-R1 showcases its understanding by first identifying the core action (\texttt{crowning}) and then methodically grounding each associated semantic role -- Agent (\texttt{woman}), Object (\texttt{tiara}), and Recipient (\texttt{woman}) -- to its corresponding entity and bounding box. This structured analysis underscores Relation-R1's ability to parse complex events involving multiple participants and their distinct roles within the scene.

\noindent\underline{\textbf{Synonymous Visualization.}}
Figure~\ref{fig:syn} displays synonymous results observed in \textit{N}-ary relation detection. As evinced by these results, our progressive CoT-guided training paradigm empowers the model to generate diverse synonymous expressions, encompassing entities, semantic roles, and activities. For instance, in Figure~\ref{fig:syn}(top), Relation-R1 demonstrates the ability to output synonymous agents such as ``\texttt{person}'' and ``\texttt{man}''. Furthermore, the model can generate semantically similar and plausible activities, exemplified by ``\texttt{barbecues}'' and ``\texttt{grills}'' (\cf Figure~\ref{fig:syn}(bottom)). Even within the same activity ``\texttt{rests}'', it can differentiate and utilize varying semantic role names like ``\texttt{item}'' and ``\texttt{bodypart}'' (\cf Figure~\ref{fig:syn}(row 2)). These synonymous outputs indicate a more flexible and robust semantic understanding beyond rigid lexical matching, thereby significantly enhancing the model's generalization capabilities to diverse linguistic formulations and real-world variations.

\section{Discussions} 
\label{supp_sec:limits}
\textbf{Limitations.}
While Relation-R1 demonstrates state-of-the-art performance in visual relation understanding, several limitations warrant discussion and offer avenues for future research. The first one is computational cost. Like other GRPO-based frameworks, the proposed framework introduces a multi-stage training process that requires substantial time to collect a training batch. The second one is the interpretability of GRPO refinements. While CoT aims to make the initial reasoning process more transparent, the subsequent policy optimization by GRPO, while guided by multi-rewards, might still operate as a ``black box'' to some extent.

\textbf{Broader Impacts.}
The development of Relation-R1 promises significant positive impacts, such as more reliable AI systems with reduced hallucinations for applications in robotics and content analysis, enhanced accessibility tools for the visually impaired through richer scene descriptions, and advanced information retrieval based on complex multi-entity interactions. However, ethical considerations include the potential misuse for generating sophisticated misinformation, the risk of amplifying societal biases present in training data, and heightened surveillance capabilities if deployed without oversight.

\clearpage
\end{document}